%% file: arxiv.tex
\documentclass[10pt,twocolumn,letterpaper]{article}

\usepackage{iccv}
\usepackage{times}
\usepackage{epsfig}
\usepackage{graphicx}
\usepackage{amsmath}
\usepackage{amssymb}

\usepackage{graphicx}
\usepackage{amsmath}
\usepackage{amssymb}
\usepackage{booktabs}
\usepackage{tikz}
\usepackage{bbm}
\usepackage{cite}
\usepackage{bm}
\usepackage{pifont}
\usepackage{cuted}
\usepackage{colortbl}
\usepackage[accsupp]{axessibility}
\usepackage[caption=false]{subfig}
\newcommand{\cmark}{\ding{51}}
\newcommand{\xmark}{\ding{55}}
\DeclareMathOperator*{\argmax}{arg\,max}

\usepackage{multirow}
\usepackage{lipsum}

\definecolor{Gray}{rgb}{0.9,0.9,0.9}
\newcolumntype{g}{>{\columncolor{Gray}}c}

\usepackage[labelfont=bf,font=small,skip=3pt]{caption}
\usepackage[pagebackref,breaklinks,colorlinks,citecolor=blue]{hyperref}

\usepackage[capitalize]{cleveref}
\crefname{section}{Sec.}{Secs.}
\Crefname{section}{Section}{Sections}
\Crefname{table}{Table}{Tables}
\crefname{table}{Tab.}{Tabs.}

\newcommand{\customfootnotetext}[2]{{
  \renewcommand{\thefootnote}{#1}
  \footnotetext[0]{#2}}}

\iccvfinalcopy

\def\iccvPaperID{9489}

\begin{document}

\setlength{\abovedisplayskip}{3pt}
\setlength{\belowdisplayskip}{3pt}
\setlength{\abovedisplayshortskip}{3pt}
\setlength{\belowdisplayshortskip}{3pt}

\include{macros}

\title{ContactGen: Generative Contact Modeling for Grasp Generation}

\author{
Shaowei Liu$^{1\dagger}$\hspace{4mm}
Yang Zhou$^2$\hspace{4mm}
Jimei Yang$^2$\hspace{4mm}
Saurabh Gupta$^{1*}$\hspace{4mm}
Shenlong Wang$^{1*}$\hspace{4mm} \\
$^1$University of Illinois Urbana-Champaign\hspace{4mm}
$^2$Adobe Research\\
{\normalsize \url{https://stevenlsw.github.io/contactgen/}}
}
\maketitle

\begin{strip}
 \vspace{-2mm}
\centering
\includegraphics[width=0.9\textwidth]{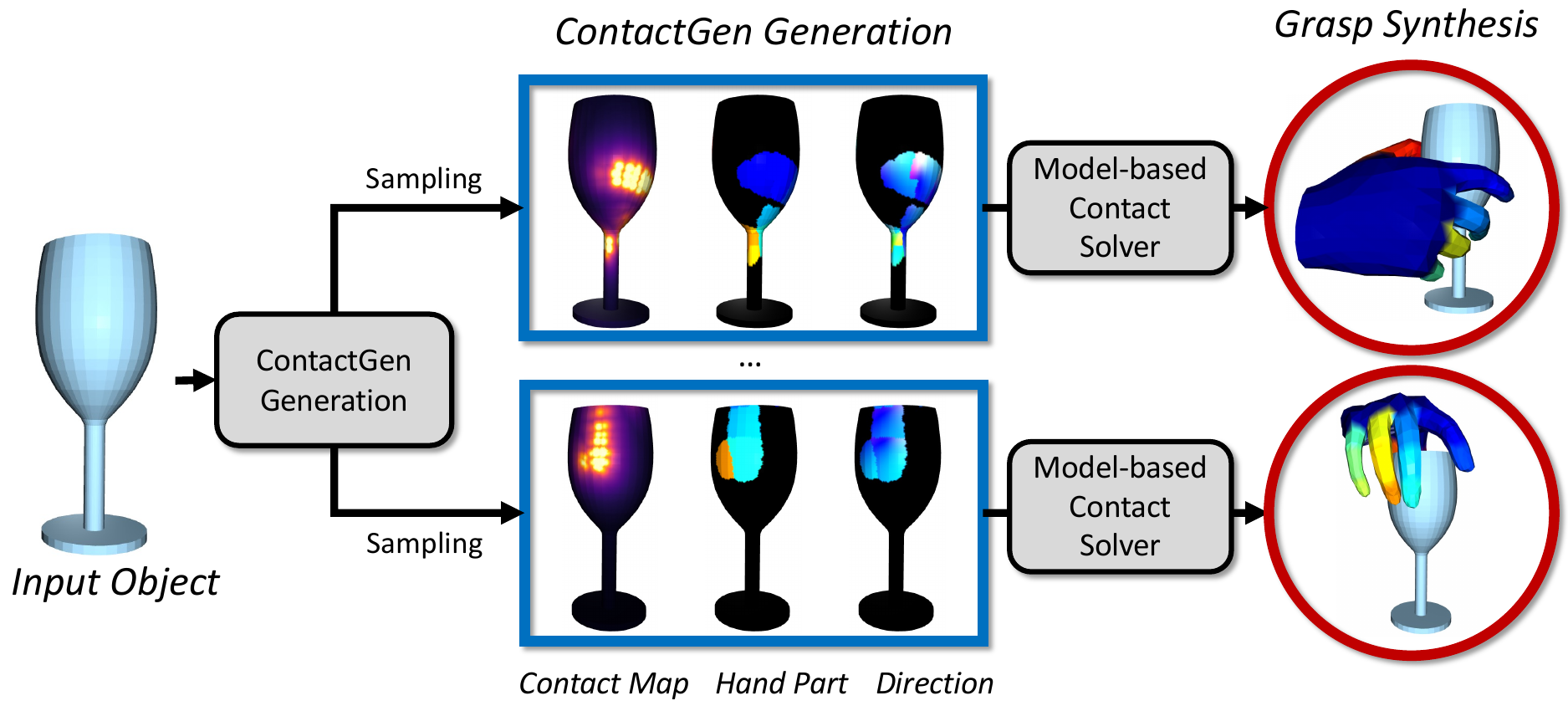}
\captionof{figure}{We introduce novel, concise and comprehensive contact representation for hand-object interaction (indicated by the dashed box). We demonstrate the process of generating contact representations from a given input object (on the left) and inferring the underlying grasp from the contact representation via model-based optimization (on the right). 
}
\label{fig: teaser}
\vspace{-4mm}
\end{strip}

\customfootnotetext{$\dagger$}{Work started at an internship at Adobe Research.}
\customfootnotetext{*}{Equal advising, alphabetic order}

\input{0_abstraction}
\input{1_intro}
\input{2_related}
\input{3_method}
\input{4_exp}
\input{5_conclusion}

{\small
\bibliographystyle{ieee_fullname}
\bibliography{biblioShort, egbib}
}

\appendix
\clearpage
\input{6_supp}

\end{document}

%% file: macros.tex
\newcommand{\tree}{\bm{\Gamma}}
\newcommand{\shenlong}[1]{\textcolor{magenta}{#1}}
\newcommand{\todocite}[1]{\textcolor{red}{\textit{Citation needed []}}}
\newcommand{\shenlongsay}[1]{\textcolor{blue}{[Shenlong: #1]}}
\newcommand{\sg}[1]{\textcolor{blue}{[Saurabh: #1]}}
\newcommand{\confirm}[1]{\textcolor{red}{#1}}
\newcommand{\todo}[1]{\textcolor{red}{\textit{TODO: #1}}}
\newcommand{\shaowei}[1]{\textcolor{magenta}{[Shaowei: #1]}}
\newcommand{\jimei}[1]{\textcolor{orange}{[Jimei: #1]}}
\newcommand{\yang}[1]{\textcolor{brown}{[Yang: #1]}}

\newcommand{\imgtile}[2]{
    {\tikz{
    \node[draw=black, draw opacity=1.0, line width=.3mm, fill opacity=0.7,fill=white, inner sep=0pt](gt) at (0, 0) {\includegraphics[width=#2\linewidth]{#1}};
    \node[draw=black, draw opacity=1.0, line width=.3mm, fill opacity=0.7,fill=white, inner sep=0pt](gt) at (-0.1, 0.1) {\includegraphics[width=#2\linewidth]{#1}};
    \node[draw=black, draw opacity=1.0, line width=.3mm, fill opacity=0.7,fill=white, inner sep=0pt](gt) at (-0.2, 0.2) {\includegraphics[width=#2\linewidth]{#1}};
    \node[draw=black, draw opacity=1.0, line width=.3mm, fill opacity=0.7,fill=white, inner sep=0pt](gt) at (-0.3, 0.3) {\includegraphics[width=#2\linewidth]{#1}}; }}
}


\newcommand{\robotD}[0]{RoboArt\xspace}
\newcommand{\sapiensD}[0]{Sapiens\xspace}
\newcommand{\wim}[0]{WatchItMove\xspace}
\newcommand{\mbs}[0]{MultiBodySync\xspace}
\newcommand{\xpar}[1]{\noindent\textbf{#1}\ \ }
\newcommand{\vpar}[1]{\vspace{3mm}\noindent\textbf{#1}\ \ }

\newcommand{\sect}[1]{Section~\ref{#1}}
\newcommand{\sects}[1]{Sections~\ref{#1}}
\newcommand{\eqn}[1]{Equation~\ref{#1}}
\newcommand{\eqns}[1]{Equations~\ref{#1}}
\newcommand{\fig}[1]{Figure~\ref{#1}}
\newcommand{\figs}[1]{Figures~\ref{#1}}
\newcommand{\tab}[1]{Table~\ref{#1}}
\newcommand{\tabs}[1]{Tables~\ref{#1}}

\newcommand{\ignorethis}[1]{}
\newcommand{\norm}[1]{\lVert#1\rVert}
\newcommand{\fcseven}{$\mbox{fc}_7$}

\renewcommand*{\thefootnote}{\fnsymbol{footnote}}

\def\naive{na\"{\i}ve\xspace}
\def\Naive{Na\"{\i}ve\xspace}

\makeatletter
\DeclareRobustCommand\onedot{\futurelet\@let@token\@onedot}
\def\@onedot{\ifx\@let@token.\else.\null\fi\xspace}

\def\iid{\emph{i.i.d}\onedot}
\def\eg{\emph{e.g}\onedot} \def\Eg{\emph{E.g}\onedot}
\def\ie{\emph{i.e}\onedot} \def\Ie{\emph{I.e}\onedot}
\def\cf{\emph{c.f}\onedot} \def\Cf{\emph{C.f}\onedot}
\def\etc{\emph{etc}\onedot} \def\vs{\emph{vs}\onedot}
\def\wrt{w.r.t\onedot} \def\dof{d.o.f\onedot}
\def\etal{\emph{et al}\onedot}
\makeatother

\definecolor{citecolor}{RGB}{34,139,34}
\definecolor{mydarkblue}{rgb}{0,0.08,1}
\definecolor{mydarkgreen}{rgb}{0.02,0.6,0.02}
\definecolor{mydarkred}{rgb}{0.8,0.02,0.02}
\definecolor{mydarkorange}{rgb}{0.40,0.2,0.02}
\definecolor{mypurple}{RGB}{111,0,255}
\definecolor{myred}{rgb}{1.0,0.0,0.0}
\definecolor{mygold}{rgb}{0.75,0.6,0.12}
\definecolor{myblue}{rgb}{0,0.2,0.8}
\definecolor{mydarkgray}{rgb}{0.66,0.66,0.66}

\newcommand{\myparagraph}[1]{\vspace{-6pt}\paragraph{#1}}

\newcommand{\bbR}{{\mathbb{R}}}
\newcommand{\bK}{\mathbf{K}}
\newcommand{\bX}{\mathbf{X}}
\newcommand{\bY}{\mathbf{Y}}
\newcommand{\bk}{\mathbf{k}}
\newcommand{\bx}{\mathbf{x}}
\newcommand{\by}{\mathbf{y}}
\newcommand{\bhy}{\hat{\mathbf{y}}}
\newcommand{\bty}{\tilde{\mathbf{y}}}
\newcommand{\bG}{\mathbf{G}}
\newcommand{\bI}{\mathbf{I}}
\newcommand{\bg}{\mathbf{g}}
\newcommand{\bS}{\mathbf{S}}
\newcommand{\bs}{\mathbf{s}}
\newcommand{\bM}{\mathbf{M}}
\newcommand{\bw}{\mathbf{w}}
\newcommand{\eye}{\mathbf{I}}
\newcommand{\bU}{\mathbf{U}}
\newcommand{\bV}{\mathbf{V}}
\newcommand{\bW}{\mathbf{W}}
\newcommand{\bn}{\mathbf{n}}
\newcommand{\bv}{\mathbf{v}}
\newcommand{\bq}{\mathbf{q}}
\newcommand{\bR}{\mathbf{R}}
\newcommand{\bi}{\mathbf{i}}
\newcommand{\bj}{\mathbf{j}}
\newcommand{\bp}{\mathbf{p}}
\newcommand{\bt}{\mathbf{t}}
\newcommand{\bJ}{\mathbf{J}}
\newcommand{\bu}{\mathbf{u}}
\newcommand{\bB}{\mathbf{B}}
\newcommand{\bD}{\mathbf{D}}
\newcommand{\bz}{\mathbf{z}}
\newcommand{\bP}{\mathbf{P}}
\newcommand{\bC}{\mathbf{C}}
\newcommand{\bA}{\mathbf{A}}
\newcommand{\bZ}{\mathbf{Z}}
\newcommand{\bff}{\mathbf{f}}
\newcommand{\bF}{\mathbf{F}}
\newcommand{\bo}{\mathbf{o}}
\newcommand{\bc}{\mathbf{c}}
\newcommand{\bT}{\mathbf{T}}
\newcommand{\bQ}{\mathbf{Q}}
\newcommand{\bL}{\mathbf{L}}
\newcommand{\bl}{\mathbf{l}}
\newcommand{\ba}{\mathbf{a}}
\newcommand{\bE}{\mathbf{E}}
\newcommand{\bH}{\mathbf{H}}
\newcommand{\bd}{\mathbf{d}}
\newcommand{\br}{\mathbf{r}}
\newcommand{\bb}{\mathbf{b}}
\newcommand{\bh}{\mathbf{h}}

\newcommand{\btheta}{\bm{\theta}}

\newcommand{\bhh}{\hat{\mathbf{h}}}
\newcommand{\ci}{{\cal I}}
\newcommand{\ct}{{\cal T}}
\newcommand{\co}{{\cal O}}
\newcommand{\ck}{{\cal K}}
\newcommand{\cu}{{\cal U}}
\newcommand{\cS}{{\cal S}}
\newcommand{\cQ}{{\cal Q}}
\newcommand{\cT}{{\cal S}}
\newcommand{\cC}{{\cal C}}
\newcommand{\cE}{{\cal E}}
\newcommand{\cF}{{\cal F}}
\newcommand{\cL}{{\cal L}}
\newcommand{\X}{{\cal{X}}}
\newcommand{\Y}{{\cal Y}}
\newcommand{\cH}{{\cal H}}
\newcommand{\cP}{{\cal P}}
\newcommand{\cN}{{\cal N}}
\newcommand{\cU}{{\cal U}}
\newcommand{\cV}{{\cal V}}
\newcommand{\cX}{{\cal X}}
\newcommand{\cY}{{\cal Y}}
\newcommand{\graph}{{\cal H}}
\newcommand{\bayes}{{\cal B}}
\newcommand{\cx}{{\cal X}}
\newcommand{\cg}{{\cal G}}
\newcommand{\cm}{{\cal M}}
\newcommand{\cM}{{\cal M}}
\newcommand{\cG}{{\cal G}}
\newcommand{\cR}{\cal{R}}
\newcommand{\eig}{\mathrm{eig}}

\newcommand{\D}{{\cal D}}
\newcommand{\bfp}{{\bf p}}
\newcommand{\bfd}{{\bf d}}

\newcommand{\cv}{{\cal V}}
\newcommand{\ce}{{\cal E}}
\newcommand{\cy}{{\cal Y}}
\newcommand{\cz}{{\cal Z}}
\newcommand{\cb}{{\cal B}}
\newcommand{\cq}{{\cal Q}}
\newcommand{\cd}{{\cal D}}
\newcommand{\bcf}{{\cal F}}
\newcommand{\cI}{\mathcal{I}}

\newcommand{\ut}{^{(t)}}
\newcommand{\up}{^{(t-1)}}

\newcommand{\bpi}{\boldsymbol{\pi}}
\newcommand{\bphi}{\boldsymbol{\phi}}
\newcommand{\bPhi}{\boldsymbol{\Phi}}
\newcommand{\bmu}{\boldsymbol{\mu}}
\newcommand{\bSigma}{\boldsymbol{\Sigma}}
\newcommand{\bGamma}{\boldsymbol{\Gamma}}
\newcommand{\bbeta}{\boldsymbol{\beta}}
\newcommand{\bomega}{\boldsymbol{\omega}}
\newcommand{\blambda}{\boldsymbol{\lambda}}
\newcommand{\bkappa}{\boldsymbol{\kappa}}
\newcommand{\btau}{\boldsymbol{\tau}}
\newcommand{\balpha}{\boldsymbol{\alpha}}
\def\bgamma{\boldsymbol\gamma}

\newcommand{\prox}{{\mathrm{prox}}}

\newcommand{\pardev}[2]{\frac{\partial #1}{\partial #2}}
\newcommand{\dev}[2]{\frac{d #1}{d #2}}
\newcommand{\dw}{\delta\bw}
\newcommand{\lab}{\mathcal{L}}
\newcommand{\unlab}{\mathcal{U}}
\newcommand{\ind}{1{\hskip -2.5 pt}\hbox{I}}
\newcommand{\ff}[2]{   \cf_{\prec (#1 \rightarrow #2)}}
\newcommand{\vv}[2]{   \cv_{\prec (#1 \rightarrow #2)}}
\newcommand{\dd}[2]{   \delta_{#1 \rightarrow #2}}
\newcommand{\ld}[2]{   \lambda_{#1 \rightarrow #2}}
\newcommand{\en}[2]{  \bD(#1|| #2)}
\newcommand{\ex}[3]{  \bE_{#1 \sim #2}\left[ #3\right]} 
\newcommand{\exd}[2]{  \bE_{#1 }\left[ #2\right]}

\newcommand{\se}[1]{\mathfrak{se}(#1)}
\newcommand{\SE}[1]{\mathbb{SE}(#1)}
\newcommand{\so}[1]{\mathfrak{so}(#1)}
\newcommand{\SO}[1]{\mathbb{SO}(#1)}

\newcommand{\poselow}{\xi}
\newcommand{\pose}{\bm{\poselow}}
\newcommand{\linpose}{\pose^\ell}
\newcommand{\cbpose}{\pose^c}
\newcommand{\rateparam}{v_i}
\newcommand{\bapose}{\bm{\poselow}_i}
\newcommand{\trackingpose}{\bm{\poselow}}
\newcommand{\rotlow}{\omega}
\newcommand{\rot}{\bm{\rotlow}}
\newcommand{\translow}{v}
\newcommand{\trans}{\bm{\translow}}
\newcommand{\hnorm}[1]{\left\lVert#1\right\rVert_{\gamma}}
\newcommand{\lnorm}[1]{\left\lVert#1\right\rVert}
\newcommand{\barate}{v_i}
\newcommand{\trackingrate}{v}
\newcommand{\imgpt}{\mathbf{u}_{i,k,j}}
\newcommand{\mappt}{\mathbf{X}_{j}}
\newcommand{\timet}[1]{\bar{t}_{#1}}
\newcommand{\mf}[1]{\text{MF}_{#1}}
\newcommand{\kmf}[1]{\text{KMF}_{#1}}
\newcommand{\Exp}{\text{Exp}}
\newcommand{\Log}{\text{Log}}

\newcommand{\R}{\mathbb{R}}

\newcommand{\cgen}[0]{ContactGen\xspace}
\newcommand{\grab}[0]{GRAB\xspace}
\newcommand{\interhand}[0]{InterHand2.6M\xspace}

%% file: 0_abstraction.tex
\begin{abstract}

This paper presents a novel object-centric contact representation \cgen for hand-object interaction. The \cgen comprises three components: a contact map indicates the contact location, a part map represents the contact hand part, and a direction map tells the contact direction within each part. Given an input object, we propose a conditional generative model to predict \cgen and adopt model-based optimization to predict diverse and geometrically feasible grasps. Experimental results demonstrate our method can generate high-fidelity and diverse human grasps for various objects.
\end{abstract}

%% file: 1_intro.tex
\section{Introduction}

Modeling hand-object interaction~\cite{ballan2012motion, panteleris2017back, tekin2019h+, hasson2019learning, Doosti2020HOPENetAG, karunratanakul2020grasping, liu2021semi, yang2021cpf, chen2022alignsdf} has gained substantial importance across various domains in animation, games, and augmented and virtual reality~\cite{ Ueda2003AHE, Hll2018EfficientPI, Wu2020HandPE, hurst2013gesture}. For instance, given an object, one would like to create a computational model to reason about the different ways a human hand can interact with it, e.g., how to grasp the object using a single hand. To ensure realism and authenticity in these interactions, a precise understanding of contact is crucial~\cite{brahmbhatt2019contactdb, brahmbhatt2020contactpose, brahmbhatt2019contactgrasp}. A thorough contact modeling should account for factors such as which regions of the object are likely to make contact, which parts of the hand will touch the object, the strength of the contact force, and the direction of the contact, among others. In contrast, the lack of thorough and precise modeling can result in unnatural and unrealistic interactions, such as insufficient contact or excessive penetration.

Previous approaches often rely on a contact map~\cite{jiang2021hand, grady2021contactopt, taheri2020grab, huang2022capturing, narasimhaswamy2020detecting}~applied on object point clouds, where values are bounded within the $[0, 1]$ range to indicate the contact status of the points. Nevertheless, simply modeling contact maps does not fully capture the details of contact. Specifically, even with the contact map, ambiguities remain regarding which regions of the hand are in contact and the manner of contact. Moreover, a single contact map falls short of representing the structured uncertainty inherent in hand-object interactions.

In this paper, we address the aforementioned challenges by introducing a novel contact representation named \cgen. \cgen provides a comprehensive representation that encodes the specific contact parts of both the object and hand, along with the precise touch direction. Specifically, for each point on the object's surface, \cgen models: (1) the contact probability of the point touched by the hand; (2) the specific part of the hand making contact, such as various fingertips or the palm, in the form of a categorical probability; and (3) the orientation of the touch with respect to the hand part making contact, represented as a vector on the unit sphere. \cref{fig: teaser} depicts the three components of the presented \cgen. Our approach significantly expands the traditional contact map, offering a precise and unambiguous representation of hand-object interaction.

Next, we introduce a novel, generative method that learns to infer diverse yet realistic \cgen for any given object. To account for uncertainties, we implement a hierarchical conditional Variational Autoencoder (CVAE)~\cite{sohn2015learning}. This CVAE sequentially models the contact map, the hand part map, and the direction map. When provided with a 3D object as conditional input, the CVAE initially models the underlying distribution of contact maps. From this, one can sample contact maps, and using these samples as additional conditioning variables, to infer the distribution of the hand part map and sample from the distribution. Lastly, direction maps can be generated based on the sampled contact map and hand part map. This sequential generation strategy decomposes variations within the entire space into distinct components, ensuring explicit uncertainty modeling for each component.

The proposed \cgen is applied to human grasp synthesis~\cite{pollard2005physically, li2007data, kry2006interaction, kalisiak2001grasp}, whose objective is to generate diverse physically plausible human grasps for various objects. In contrast to existing work~\cite{jiang2021hand, corona2020ganhand, karunratanakul2020grasping, taheri2020grab, karunratanakul2021skeleton}, which primarily addresses grasp uncertainty within the hand space, our key innovation lies in addressing this uncertainty within the object space. We achieve this by designing a novel contact solver that effectively derives hand grasp poses from \cgen. The \cgen is sampled from the CVAE model conditioned on specific object. As a result, our design yields more realistic and natural hand grasps, as shown by improved contact, reduced penetration, and increased stability. Additionally, the hierarchical contact modeling fosters greater diversity in the generated grasps. Our experiments validate the effectiveness of our method in ensuring both diversity and fidelity in hand-object interactions, surpassing the performance of current state-of-the-art techniques.

In summary, our contributions are as follows:
\begin{itemize}
\item We introduce \cgen, a \textbf{novel object-centric representation} that concurrently models the point-wise contact parts of both the object and the hand, as well as the contact direction relative to each part of the hand.
\item We propose a \textbf{sequential CVAE} that learns to model the uncertainty inherent in hand-object interactions using our contact representation.
\item We develop a \textbf{novel contact solver} for human grasp synthesis by integrating our proposed generative contact modeling with model-based optimization. This combination yields superior fidelity and diversity compared to existing methods.
\end{itemize}

%% file: 2_related.tex
\section{Related Work}

\begin{table}
\setlength{\tabcolsep}{4pt}
\resizebox{\linewidth}{!}{
\begin{tabular}{lccccc}
\toprule
\bf Method & \bf Hand Model & \multicolumn{3}{c}{\bf Contact Modeling} & \bf Object-centric\\
\cmidrule(lr){3-5}
  & & \bf location & \bf part & \bf direction \\
\midrule
Grasping Field~\cite{karunratanakul2020grasping} & point cloud & \cmark & \cmark & \xmark & \xmark \\
GraspTTA~\cite{jiang2021hand} & mesh & \cmark & \xmark & \xmark & \xmark \\
ContactOpt~\cite{grady2021contactopt} & mesh & \cmark & \xmark & \xmark & \xmark \\
TOCH~\cite{zhou2022toch} & mesh & \cmark & \cmark & \xmark & \cmark \\
Ours & mesh/sdf & \cmark & \cmark & \cmark & \cmark \\

\bottomrule
\end{tabular}}
\caption{Contact representation comparison between different methods of hand-object interaction. Most existing work adopt contact map, which is insufficient to recover the corresponding grasp.}
\label{table: contact_rep}
\vspace{-4mm}
\end{table}

\noindent {\bf Contact Modeling}
Various contact representations have been proposed in hand-object interaction~\cite{rijpkema1991computer, kry2006interaction, hamer2010object, ye2012synthesis, liu2009dextrous, yang2021cpf, jiang2021hand, grady2021contactopt}, and human-object interactions~\cite{xie2022chore, hassan2021populating, savva2016pigraphs, zhang2020place, zhang2022couch, bhatnagar2022behave, fieraru2020three, chen2019holistic++, huang2022capturing}.  As shown in \cref{table: contact_rep}, existing studies~\cite{jiang2021hand, grady2021contactopt, taheri2022goal, wu2022saga, tse2022s} primarily adopt the contact map as a standard representation. Zhou \etal~\cite{zhou2022toch} proposed an object-centric TOCH field by encoding contact locations and hand correspondences on the object aimed for temporal hand pose denoising. Nonetheless, we argue the information provided by contact maps~\cite{grady2021contactopt} or sparse hand-object correspondences on contact locations~\cite{zhou2022toch} falls short. Contact maps lack information about their counterparts, while sparse correspondences cannot provide detailed contact directions. In our approach, we not only infer the contact location, but also the hand part in contact and its touch direction.
This novel representation is point-wise and object-centric, which does not require any input from the hand, but enables the comprehensive decoding of hand information.

\noindent {\bf Grasp Synthesis}
Grasp synthesis has gained extensive attention across both robotic hand manipulation~\cite{antotsiou2018task, mahler2019learning, brahmbhatt2019contactgrasp, hsiao2006imitation, miller2004graspit}, animation~\cite{elkoura2003handrix, borst2005realistic, rijpkema1991computer, kalisiak2001grasp}, digital human synthesis~\cite{sahbani2012overview, zhang2021manipnet, li2007data, kry2006interaction}, and physical motion control~\cite{kim2015physics, pollard2005physically, holden2017phase}. In this work, we focus on realistic human grasp synthesis~\cite{taheri2020grab, jiang2021hand, karunratanakul2021skeleton, karunratanakul2020grasping,  corona2020ganhand}. The objective is to generate authentic human grasps of diverse objects. The key challenge is to achieve both physical plausibility and diversity within the generated grasps. A majority of existing approaches employ CVAE to sample hand MANO parameters~\cite{taheri2020grab, jiang2021hand, taheri2022goal, tendulkar2023flex} or hand joints~\cite{karunratanakul2021skeleton}, which primarily model grasp variations within the hand space. These model tends to easily overfit to common grasp patterns, lacking diversity despite the use of CVAE. Karunratanakul \etal~\cite{karunratanakul2020grasping} proposed to learn an implicit grasping field. However, this approach does not account for hand articulations; instead, it treats posed hands as rigid objects, causing the solution space to span the entire configuration space and the generated results do not guarantee to be valid. In contrast to existing methods, we suggest learning the object-centric \cgen within the object space. This involves breaking down the variability in hand grasping into distinct components within the \cgen: contact location, hand part, and touch direction. The decomposition allows us to generate physically realistic grasps with increased diversity by sampling from \cgen, while existing approaches model grasp uncertainty in the hand space often lean towards learning generalized grasp patterns with limited variety.

\noindent {\bf Grasp Optimization}
Another distinct research direction focuses on analytical grasp solution~\cite{krug2010efficient, seo2012planar, depierre2018jacquard, mousavian20196, miller2004graspit, turpin2022grasp, antotsiou2018task, kim2015physics, tzionas2016capturing, hasson2019learning, yang2021cpf}. The objective is to optimize grasps to minimize penetration, enhance contact, and improve overall stability. 
In human grasp generation, prior research~\cite{jiang2021hand, zhou2022toch, grady2021contactopt} optimized MANO model~\cite{romero2017embodied} parameters under contact loss and penetration loss. However, the optimization proves challenging. First, the objectives of promoting contact and reducing penetration inherently conflict with one another. Striking a balance between encouraging hand-object contact while preventing penetration is computationally intricate. Second, as shown in previous work~\cite{deng2020nasa, mihajlovic2021leap}, the inherent discretization and limited spatial resolution of mesh structures pose constraints. To address these challenges, we propose a hand articulation model that employs part-wise Signed Distance Function (SDF) for optimization. The SDF neatly partition the space for contact ($\text{SDF}=0$) and penetration ($\text{SDF}<0$). It captures fine-grained deformation and supports querying contact direction within each part, making it seamlessly compatible with our contact representations. The piecewise model also shares the same parameters as MANO. By incorporating the piecewise hand model into our optimization, we substantially enhance the grasp quality, leading to more realistic and diverse grasp results.

%% file: 3_method.tex
\section{Overview}

\begin{figure}[t]
\centering
     \includegraphics[width=0.85\linewidth]{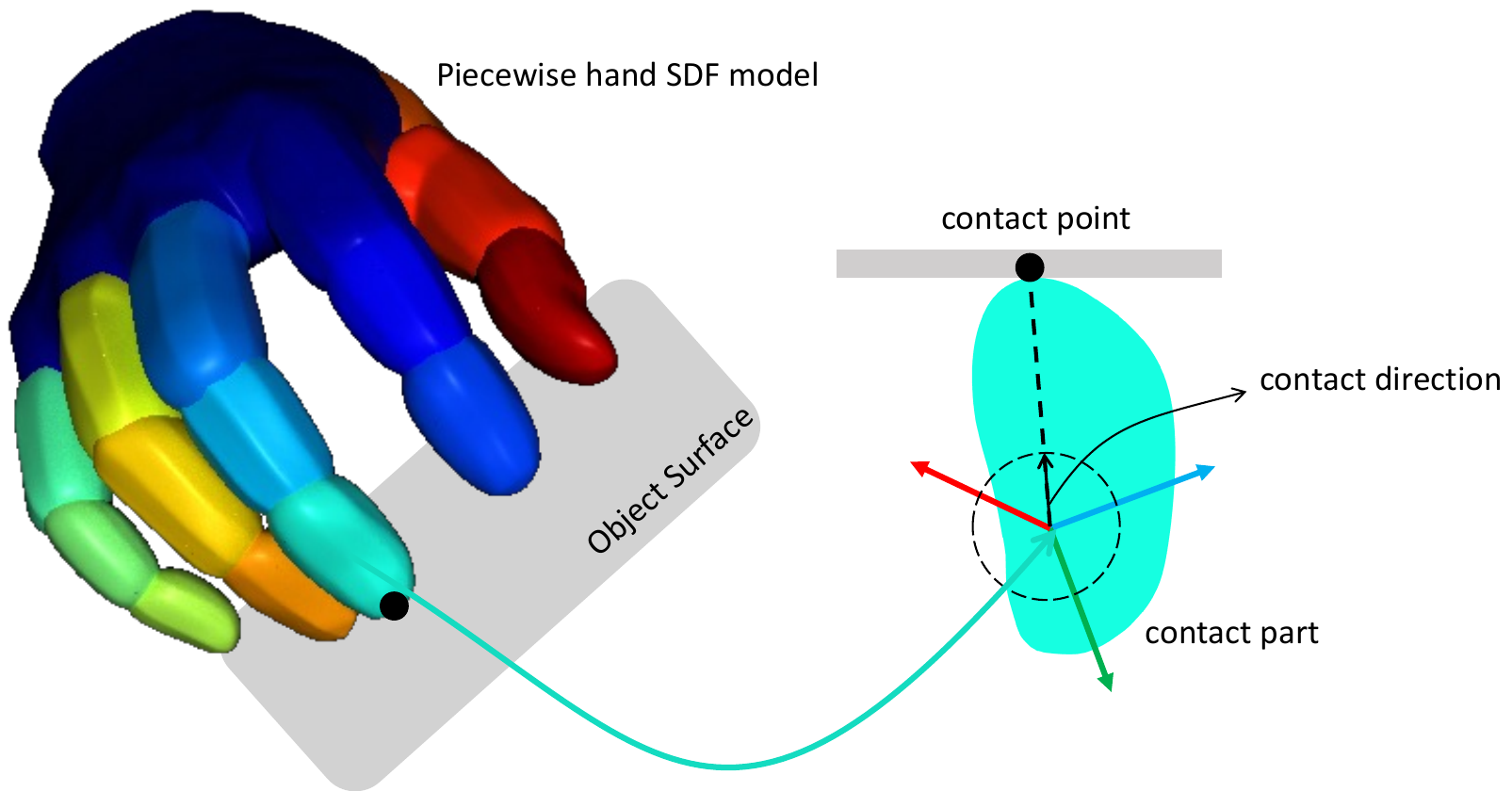} 
    \caption{Illustration of the contact part and contact direction in \cgen representation. The hand is partitioned into $B$=16 parts, represented in different color. The contact part of the point is taken as the closest hand part label. The contact direction is a unit vector from part center towards the contact point.}
    \vspace{-2mm}
    \label{fig: hand_sdf}
\end{figure}

Our technical sections are organized as follows: In \cref{sec: contact_field}, we introduce the proposed contact representation. Our novel representation encodes contact information into three maps: the contact probability map, the hand part map, and the direction map. In \cref{sec: learning}, we use a sequential CVAE, trained on hand-object interaction data, to infer the contact representation for a given object in a generative manner. In \cref{sec: grasp}, we decode the grasp from the predicted \cgen using our specially-developed contact solver, which integrates the piecewise hand articulation model. ~\cref{fig: teaser} provides a summary of our approach.

\section{Object-Centric Contact Representation}
\label{sec: contact_field}

Our object-centric contact representation $\bF = (\bC, \bP, \bD)$ consists of three maps, contact map $\bC$, part map $\bP$ and direction map $\bD$. All maps are defined on a set of $N$ points $\mathbf{O} \in \R^{N \times 3}$ sampled from the object surface, as shown in \cref{fig: teaser}. 

\vspace{-1em}
\paragraph{Contact Map} 
The contact map $\bC \in \R^{N \times 1}$, each $c_i \in \bC$ is within [0, 1], 
representing the contact probability of the point. The contact definition and computation follows~\cite{grady2021contactopt}. 
Intuitively, the contact map illustrates {\it which part of the object will likely be contacted by hand}. However, relying solely on contact maps is insufficient for complex human-object interaction modeling due to ambiguities regarding {\it how and where the hand touches the objects}. To address this, our object-centric representation is extended by explicitly modeling the other two maps.

\vspace{-1em}
\paragraph{Part Map} To locate the in-contact point on the hand surface, we use a part map $\bP \in \R^{N \times B}$ (one-hot vector) to indicate the hand part label in $\{1, \cdots, B\}$ in contact with the object point $\mathbf{O}$. The hand is divided into $B$ parts, the partition is shown in \cref{fig: hand_sdf}. Each value $\bp_i$ in $\bP$ is taken as the closest hand part label.

\vspace{-1em}
\paragraph{Direction Map}  Within each part, to describe an arbitrary point exactly on the part surface, we use its direction to the part center. The direction map $\bD \in \R^{N \times 3}$, $\bd_i\in \bD$ records the direction of this point w.r.t. part $b$, as shown in \cref{fig: hand_sdf}. Imagine each part as a unit sphere, the contact direction could be any ray shooting from the part center to the sphere surface. Given the direction $\bd_i$, the contact point location in part $b$ could be uniquely determined by searching along the ray direction $\bd_i$ until its part $\text{SDF}=0$.

\begin{figure}[t]
    \centering
    \includegraphics[width=\linewidth]{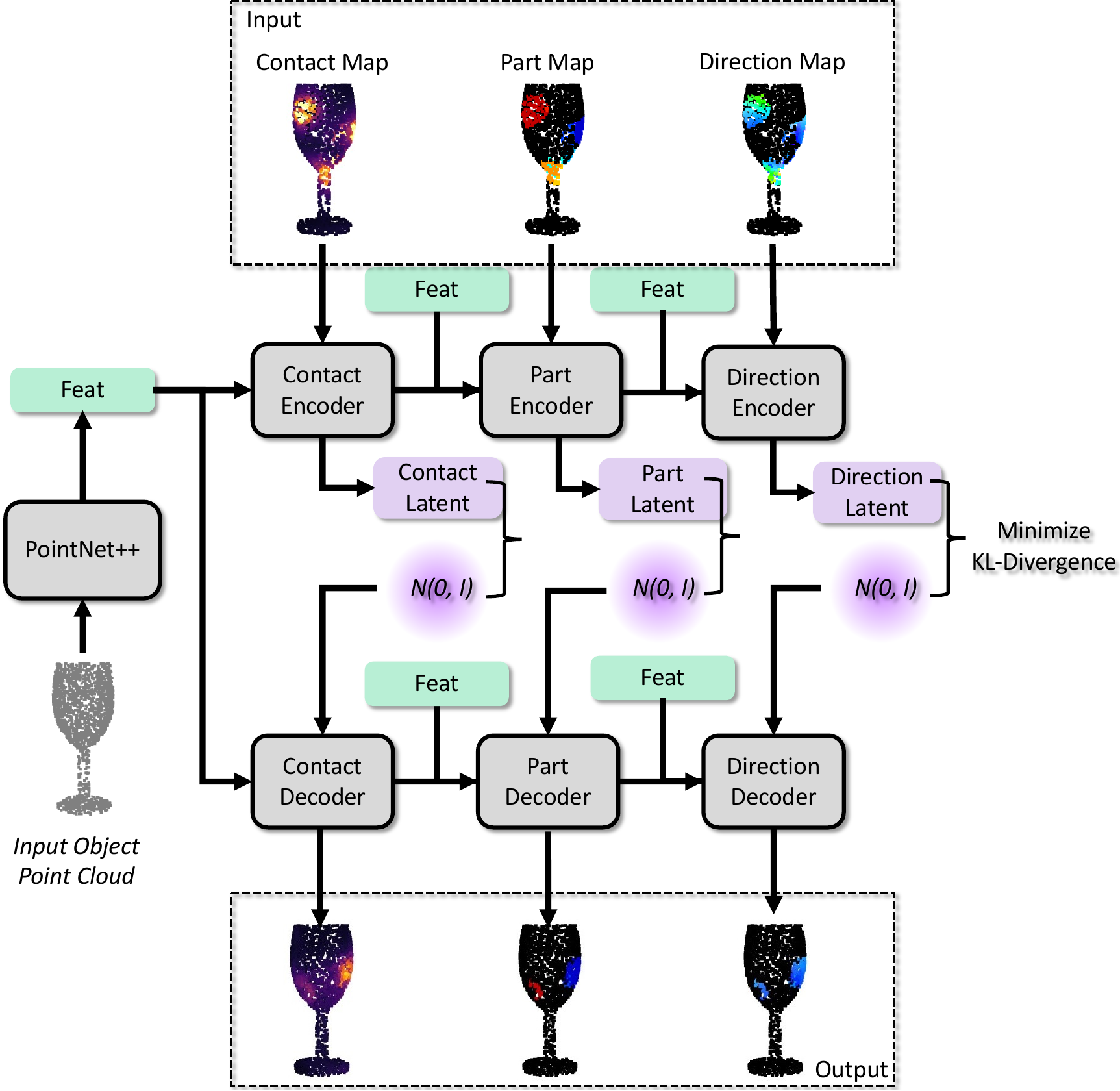}
    \caption{\cgen CVAE model architecture. Conditioned on the input object point cloud, we decompose the \cgen into individual components by a sequential encoder-decoder.}
    \label{fig: cvae}
    \vspace{-4mm}
\end{figure}

\section{Generative Contact Modeling}
\label{sec: learning}
Given sampled object point clouds $\mathbf{O}$ as input, we use a conditional generative framework to infer possible object-centric contact representations $\bF$ from the underlying distribution $p(\bF|\mathbf{O})$. 

\vspace{-3mm}
\paragraph{Sequential CVAE.} 
As shown in \cref{fig: cvae}, we model \(p(\mathbf{F}|\mathbf{O})\) sequentially using a CVAE.
We choose CVAE for its simplicity and capability to model multi-modal uncertainty. 
We factorize the joint distribution of the \cgen \(\mathbf{F} = (\mathbf{C}, \mathbf{P}, \mathbf{D})\) into a product of three conditional probabilities:
\begin{equation}
\label{eq: part decompose}
    p(\bF|\mathbf{O}) = p(\bD|\bP, \mathbf{O})p(\bP| \bC, \mathbf{O})p(\bC|\mathbf{O})
\end{equation}
The contact map $\bC$ is conditioned on object input $\mathbf{O}$; the part map $\bP$ is additionally conditioned on contact map $\bC$; direction map additionally conditioned on part map $\bP$. We control each component in \cref{eq: part decompose} by a latent code $z$ sampled from Gaussian distribution. We sample latent code from posterior $\bz_c \sim \mathcal{N} (\bm{\mu}_c, \mathbf{\Sigma}_c)$, $\bz_p \sim \mathcal{N} (\bm{\mu}_p, \mathbf{\Sigma}_p)$, $\bz_d \sim \mathcal{N} (\bm{\mu}_d, \mathbf{\Sigma}_d)$ during training, and sample latent code from prior $\bz_c \sim \mathcal{N} (\bm{0}, \mathbf{I})$, $\bz_p \sim \mathcal{N} (\bm{0},  \mathbf{I})$, $\bz_d \sim \mathcal{N} (\bm{0}, \mathbf{I})$ at inference. $\mathcal{G}_c$, $\mathcal{G}_p$ and $\mathcal{G}_d$ denote the conditional decoder for contact map, part map and direction map. Thus, we could sample prediction $\hat{\bC}, \hat{\bP}, \hat{\bD}$ using:
\begin{align}
     \mathbf{\hat{C}} = \mathcal{G}_c(\bz_c; \mathbf{O}), 
     \mathbf{\hat{P}} = \mathcal{G}_p(\bz_p; \mathbf{\hat{C}}; \mathbf{O}), 
     \mathbf{\hat{D}} = \mathcal{G}_d(\bz_d; \mathbf{\hat{P}}, \mathbf{O})
\end{align}
The sequential structure guarantees the three generated maps are consistent with each other and decompose the complicated contact sampling into the conditional generation of each component. We could recover the complete hand information from the sampled representation $\hat{\bC}, \hat{\bP}, \hat{\bD}$.

\vspace{-3mm}
\paragraph{Model architecture} Our generative model is a point-based network that operates on the sampled point cloud of the input object. We extract shared object features using PointNet++~\cite{qi2017pointnet++}. The Gaussian parameters of each component are inferred by an encoder modeled as a MLP. After sampled latent code, we have three sequential PointNet~\cite{qi2017pointnet} decoders $\mathcal{G}_c$, $\mathcal{G}_p$ and $\mathcal{G}_d$ to decode each map. For part information, we employ an embedding layer to encode each part label into an embedded space before feeding it to the network. The detailed architecture is shown in the supplementary.

\vspace{-3mm}
\paragraph{Training} We train the network in an end-to-end fashion. All networks are trained jointly. We use teacher forcing~\cite{williams1989learning} by sending GT contact map and part map as conditioning during training. The VAE loss consists of a reconstruction term and a KL regularization term. The total loss $\mathcal{L}$ is the following:
\begin{equation}
    \mathcal{L} = \mathcal{L}_{rec} + \lambda_{KL}\mathcal{L}_{KL}
\end{equation}
The reconstruction $\mathcal{L}_{rec}$ is defined as:
\begin{equation}
     \mathcal{L}_{rec} = W_C \left(|C - \hat{C}| + \lambda_p\mathcal{L}_{CE}(P, \hat{P}) + \lambda_{d}\mathcal{L}_d(D, \hat{D})\right)
\end{equation}
$\mathcal{L}_{CE}$ is the standard cross-entropy loss between the predicted part label $\hat{P}$ and ground truth (GT) $P$, $\mathcal{L}_d$ computes the cosine similarity between the GT direction $\mathbf{d}_i$ and predicted direction $\hat{\mathbf{d}_i}$ for each point. All losses are computed per-point and weighted by $W_C= C + \delta$, where $C \in [0, 1]$ is the GT contact map value as we focus more on those contact locations and $\delta$ is a default weight for non-contacted points. $\mathcal{L}_{KL}$ regularizes the latent z-space close to normal distribution $\mathcal{N}(\mathbf{0}, \mathbf{I})$, $\mathcal{L}_{KL}(\bm{\mu}, \bm{\Sigma})=KL[\mathcal{N}(\bm{\mu}, \bm{\Sigma}^2)||\mathcal{N}(\mathbf{0}, \mathbf{I})]$. The whole KL loss consists KL regularization from each latent space $\mathcal{L}_{KL} = \mathcal{L}_{KL}(\bm{\mu}_c, \bm{\Sigma}_c) + \mathcal{L}_{KL}(\bm{\mu}_p. \bm{\Sigma}_p) + \mathcal{L}_{KL}(\bm{\mu}_d, \bm{\Sigma}_d)$.

\section{Grasp Synthesis}
\label{sec: grasp}
Next, we discuss how to convert the sampled object-centric contact representation into the corresponding articulated hand grasp. To achieve this, we parameterize the hand using the piecewise SDF model, formulate a model-based optimization to recover the most plausible hand pose.

\subsection{Piecewise hand articulation model}
Following~\cite{lombardi2021latenthuman}, we convert MANO model ~\cite{romero2017embodied} to a piecewise SDF model. This modification enhances the compatibility of the hand model with the proposed \cgen for grasp synthesis. The piecewise model partitions the hand into $B$ parts and use piecewise SDF $\{\text{SDF}_b\}_{b=1}^{B}$ to represent each part. The model is parameterized by part pose $\bT_b \in \SE3$ and a global shape vector $\mathbf{\beta}$, $\bT_b$ is the transformation from part's local coordinate frame to the global. We use axis-angle $\theta_b$ as pose code for each part, $\bT_b = \bT(\theta_b)$. Given part $b$, the signed distance from a point $\bo_i$ to its surface is given by $\text{SDF}_b(\bf{T}_b^{-1}\bo_i; \beta) = \text{SDF}_b({\bf{T}(\theta_b)}^{-1}\bo_i; \beta)$.  The direction of the point w.r.t the part is given by $\bd_i = \frac{\bf{T}_b^{-1}\bo_i}{\|\bf{T}_b^{-1}\bo_i\|}$. The overall piecewise hand SDF model parameters are pose parameters concatenated from all parts $\theta = \oplus_{b}^{B}\theta_b \in \R^{B \times 3}$ and shape code $\beta$. The parameters $\theta$ and $\beta$ are shared across the MANO model~\cite{romero2017embodied} and the piecewise SDF hand model. One can easily convert between each other. Training details of the SDF model are provided in the supplementary.

\subsection{Contact solver}
Given the sampled points $\mathbf{O}$ and predicted \cgen $\hat{\bC} \in \R^{N \times 1}$, $\hat{\bP} \in \R^{N \times B}$ in one-hot format, $\hat{\bD} \in \R^{N \times 3}$, the goal is to infer the hand model $\{\text{SDF}_b\}_{b=1}^{B}$ parameters $\theta$ and $\beta$. The optimization objective is the following:

\begin{equation}
\label{eq: grasp_opt}
    \min_{\theta, \beta} \lambda_c\mathcal{L}_c + \lambda_d \mathcal{L}_d + \lambda_{p} \mathcal{L}_p + \lambda_{r} \mathcal{L}_r
\end{equation}
The $\mathcal{L}_c$ denotes the contact map loss, Given object points $\mathbf{O}$, For each part $b$ we compute $\text{SDF}_b(\bT(\theta_b)^{-1}\mathbf{O};\beta) \in \R^{N \times 1}$, $\hat{\bP}_b$ denotes the b-th column of $\hat{\bP}$.
The contact map loss is: 
\begin{equation}
\mathcal{L}_c = \hat{\bC} \sum_{b=1}^B \hat{\bP}_b \cdot |\text{SDF}_b|
\end{equation}
For $\bo_i$ with higher contact value $\hat{c}_i$ and predicted part $\argmax\hat{p}_i=b$, we encourage the SDF of the point in hand part $b$ to be close to 0, driving the hand to touch the contact location. The second term $\mathcal{L}_d(\bD, \mathbf{\hat{D}})$ encourages the point direction $\bd(\bo_i, \theta_b)$ of hand part $b$ to match the predicted direction $\mathbf{\hat{d_i}}$. 
\begin{equation}
\mathcal{L}_d = W_C \left(1 - \text{cos}(\bD, \mathbf{\hat{D}})\right)
\end{equation}
The penetration loss $\mathcal{L}_p$ prevents object sampled points from being inside the hand:
\begin{equation}
    \mathcal{L}_p = \sum_{b=1}^{B}-\text{max}\left(\text{SDF}_b, 0\right)
\end{equation}
The last term $\mathcal{L}_r$ is the regularization term that prevents the model from being too complex, $\mathcal{L}_r = \|\theta\|^2 + \|\beta\|^2$. 
For each object with predicted \cgen, we optimized the above objective function to get the hand grasp. 

\vspace{-2mm}
\paragraph{Inference} We optimize $\theta$ and $\beta$ from scratch. Following~\cite{zhou2022toch, wu2022saga}, we adopted a two-stage optimization strategy. In the first stage, we only optimize the global pose of the hand. In the second stage, we freeze the hand's global pose and optimize the hand's pose and shape parameters. We use Adam~\cite{kingma2014adam} optimizer for both stages.

%% file: 4_exp.tex
\section{Experiments}

\subsection{Datasets} We use the \grab dataset~\cite{taheri2020grab} to train the \cgen CVAE and test grasp synthesis performance. GRAB contains real human grasps for 51 objects from 10 different subjects. We follow the official train/test split. The test set contains six unseen objects. Following~\cite{karunratanakul2020grasping, karunratanakul2021skeleton, jiang2021hand}, we also test on out-of-domain objects from HO3D dataset~\cite{hampali2020honnotate} test set to evaluate the generalization ability of the model.

\subsection{Implementation Details}
The piecewise hand SDF model was trained on the Freihand dataset~\cite{zimmermann2019freihand} with 32,560 samples. The \cgen CVAE takes $N=2048$ points sampled from object surface as input. During training, we apply data augmentation by randomly rotating the object by $[-\frac{\pi}{6}, \frac{\pi}{6}]$ around each axis. The latent dimension of the CVAE was set to 16. We set $\lambda_{p}=0.5$, $\lambda_{d}=1$, and the KL weight $\lambda_{KL}$ was annealed from 0 to $5e-2$ during training. We employed the standard Adam optimizer \cite{kingma2014adam} with a learning rate of $1.6e-3$ and a batch size of 256. The CVAE was trained for 3000 epochs.

For grasp synthesis, we do 200 iterations with a learning rate of $5e-2$ to optimize the global hand translation and rotation, and 1000 iterations with a learning rate of $5e-3$ to optimize hand pose and shape. The regularization term $\lambda_c=1e-1$, $\lambda_d=1e-2$, $\lambda_r=1e-2$, penetration weight $\lambda_p=3.0$. We use the Adam optimizer for both stages.

\subsection{Recovering Grasps from GT \cgen}
\label{sec: eval_contact_field}

We first evaluate the effectiveness of the proposed \cgen representation and optimization procedure at recovering hand grasps from ground truth \cgen. We compare two alternate representations from past work.

\noindent {\bf Baselines.} \textbf{ContactOpt}~\cite{grady2021contactopt} utilizes contact maps on both the hand and object to refine hand pose. In our experiment, we provide it with necessary GT contact maps. \textbf{TOCH}~\cite{zhou2022toch} uses binary contact labels on the object's surface and hand correspondences on the MANO mesh~\cite{romero2017embodied} to represent contacts. For our study, we adapt TOCH to work under a single frame by removing temporal constraints. Additionally, we convert the GT contact map into binary using a threshold of 0.5 and provide the associated GT contact vertices for optimization. In all cases, the methods optimize the hand pose from scratch.

\begin{table}[t]
\footnotesize
\centering
\resizebox{\linewidth}{!}{
\begin{tabular}{lcccc}
\toprule
\multirow{2}{*}{\bf Method} & \multirow{2}{*}{\bf  EPE (cm) $\downarrow$}  & \multirow{2}{*}{\bf AUC  $\uparrow$} & \bf F-score $\uparrow$ & \bf F-score $\uparrow$ \\
 &  &  & \bf @5mm & \bf @15mm \\
\midrule
ContactOpt~\cite{grady2021contactopt} & 7.00 & 0.26 & 0.24 & 0.50  \\
TOCH~\cite{zhou2022toch} & 3.44 & 0.51 & 0.39 & 0.72 \\
Ours & \textbf{1.49} & \textbf{0.77} & \textbf{0.55} & \textbf{0.91}\\
\bottomrule
\end{tabular}}
\vspace{1em}
\caption{\textbf{Contact reconstruction comparison.} We compare against ContactOpt and TOCH 's representation ability to recover GT hand grasps from contact. Due to the completeness of our representation, we are able to effectively recover the hand pose from the contact, achieving the lowest reconstruction error.}
\label{table: result_contact}
\vspace{-2mm}
\end{table}

\input{tables/contact_compare_vis}

\noindent {\bf Metrics.} We use the same three metrics as used in past works~\cite{zimmermann2019freihand, hampali2020honnotate}. \textbf{Mesh endpoint error (EPE)} measures the average Euclidean distance between the hand vertices of the prediction and the GT. \textbf{Mesh AUC} measures the percentage of correctly reconstructed vertices (vertices within 5cm are considered correct). \textbf{Mesh F-Score} calculates the harmonic mean of recall and precision between two meshes given a distance threshold. We report F-score at 5mm and 15mm.

\noindent {\bf Results.} The experimental results are presented in~\cref{table: result_contact}. ~\cref{fig: contact_compare} shows qualitative comparison of each approach. We observe that both ContactOpt and TOCH are unable to accurately recover the GT hand pose due to the ambiguity of their respective representations. ContactOpt~\cite{grady2021contactopt}, despite having access to the GT hand and object contact maps, faces challenges in determining the specific hand-part that should establish contact with a given contact location on the object. TOCH~\cite{zhou2022toch} shows comparatively better performance due to the richer contact information from hand-object correspondences. However, the optimization remains challenging as contact location is sparse. The hand pose isn't uniquely defined since the contact direction can also vary. Due to the completeness of the representation, our method is able to effectively recover the hand pose from the contact and achieves the lowest reconstruction error.

\subsection{Grasp Synthesis Evaluation} 
\label{sec: eval_grasp}
We follow the experimental setup from~\cite{karunratanakul2021skeleton}. We test on 6 unseen objects from the \grab dataset~\cite{taheri2020grab} and out-of-domain test objects from the HO3D dataset~\cite{hampali2020honnotate}.

\noindent {\bf Baselines.} 
GrabNet~\cite{taheri2020grab} and GraspTTA~\cite{jiang2021hand} utilize CVAE to generate MANO parameters. \textbf{GraspTTA}~\cite{jiang2021hand} also employs test-time adaptation to enhance generated grasps. \textbf{Grasping Field}~\cite{karunratanakul2020grasping} (GF) uses a CVAE to predict 3D hand point clouds and fit a MANO model afterward. \textbf{HALO}~\cite{karunratanakul2021skeleton} generates 3D keypoints using a CVAE and uses an implicit occupancy network to transform these keypoints into meshes. While all baselines predict within the hand space, our approach makes inferences in the object space.

\noindent {\bf Metrics.}
Following~\cite{karunratanakul2021skeleton, jiang2021hand, karunratanakul2020grasping, taheri2020grab, wu2022saga, hasson2019learning, tendulkar2023flex}, we evaluate the generated grasps based on their a) physical plausibility and stability, b) diversity, and c) perceptual attributes. 

\noindent \textbullet~To assess physical plausibility, we use hand-object \textit{Interpenetration Volume} and \textit{Contact Ratio} following~\cite{karunratanakul2021skeleton, jiang2021hand, hasson2019learning, wu2022saga, turpin2022grasp, zhang2020place, zhang2020generating}. We compute interpenetration volume by voxelizing the meshes into $1\text{mm}^3$ cubes and measuring overlapping voxels. Contact ratio calculates the proportion of grasps that are in contact with objects. 
For grasp stability assessment, consistent with~\cite{tzionas2016capturing, jiang2021hand, corona2020ganhand, hasson2019learning, turpin2022grasp, karunratanakul2020grasping}, we place the object and the predicted grasp into a simulator~\cite{coumans2013bullet}, and measure the average \textit{Simulation Displacement} of the object's center of mass under the influence of gravity. 

\noindent \textbullet~Following~\cite{karunratanakul2021skeleton, zhang2020generating}, we evaluate diversity in generated grasps by first clustering generated grasps into 20 clusters using K-means and then measuring the \textit{Entropy} of cluster assignments and the average \textit{Cluster Size}. Higher entropy and cluster size values indicate better diversity. Following previous work~\cite{karunratanakul2021skeleton}, we perform K-means clustering on 3D hand keypoints for all methods. 

\noindent \textbullet~Following~\cite{jiang2021hand, wu2022saga, karunratanakul2020grasping, tendulkar2023flex}, we also perform \textit{Human Evaluation} on the naturalness and stability of generated grasps.

\begin{table}[t]
\setlength{\tabcolsep}{2.5pt}
\renewcommand{\arraystretch}{1.05}
\footnotesize
\centering
\resizebox{\linewidth}{!}{
\begin{tabular}{lcccgg}
\toprule
 &  \bf Penetration & \bf Contact & \bf Simulation & \bf Entropy & \bf Cluster \\
       \bf Method  &  \bf Volume $\downarrow$ & \bf Ratio $\uparrow$ & \bf Displacement $\downarrow$ & \bf $\uparrow$ & \bf Size $\uparrow$\\
\midrule
GrabNet~\cite{taheri2020grab} & 3.65 & \textbf{\underline{0.96}} & \textbf{1.72} & 2.72 & 1.93 \\
HALO~\cite{karunratanakul2021skeleton} & 3.61 & 0.94 & 2.09 & \textbf{\underline{2.88}} & 2.15\\
Ours & \textbf{2.72} & \textbf{\underline{0.96}} & 2.16 & \textbf{\underline{2.88}} & \textbf{4.11} \\
\bottomrule
\end{tabular}}
\vspace{1em}
\caption{\textbf{Grasp result on the \grab dataset~\cite{taheri2020grab}}. 
Our method achieves the lowest penetration, highest contact, and comparable stability compared to previous approaches, while showcasing significantly larger generation diversity.}
\label{table: exp_grasp}
\vspace{-3mm}
\end{table}
\renewcommand{\arraystretch}{1.0}

\begin{table}[t]
\setlength{\tabcolsep}{2.5pt}
\renewcommand{\arraystretch}{1.05}
\footnotesize
\centering
\resizebox{\linewidth}{!}{
\begin{tabular}{lcccgg}
\toprule
 &  \bf Penetration & \bf Contact & \bf Simulation & \bf Entropy & \bf Cluster \\
       \bf Method  &  \bf Volume $\downarrow$ & \bf Ratio $\uparrow$ & \bf Displacement $\downarrow$ & \bf $\uparrow$ & \bf Size $\uparrow$\\
\midrule
GraspTTA~\cite{jiang2021hand} & \textbf{7.37} & 0.76 &  5.34 & 2.70 & 1.43 \\
GrabNet~\cite{taheri2020grab} & 15.50 & 0.99 & \textbf{2.34} & 2.80 & 2.06 \\
GF~\cite{karunratanakul2020grasping} & 93.01 & \textbf{1.00} & - & 2.75 & 3.44 \\
HALO~\cite{karunratanakul2021skeleton} & 25.84 & 0.97 & 3.02 & \textbf{\underline{2.81}} & 4.87 \\
Ours & 9.96 & 0.97 & 2.70 & \textbf{\underline{2.81}} & \textbf{5.04} \\
\bottomrule
\end{tabular}}
\vspace{1em}
\caption{\textbf{Grasp result on the HO3D dataset~\cite{hampali2020honnotate}.} Our method achieves performance close to the best method in each metric, while maintaining the highest diversity in the generated grasps.}
\label{table: result_ho3d}
\vspace{-3mm}
\end{table}
\renewcommand{\arraystretch}{1.0}

\input{tables/grab_vis}
\input{tables/grab_diverse_vis}

\noindent {\bf Results on \grab dataset.}
\cref{table: exp_grasp} shows comparison of our method with GrabNet~\cite{taheri2020grab} and HALO~\cite{karunratanakul2021skeleton} on the \grab dataset. For each method, we randomly generate 20 grasps. 
Our method achieves the lowest penetration, highest contact ratio, and comparable stability compared to previous approaches. It stands out in terms of grasp variability, as indicated by the significantly larger cluster size value compared to previous methods. Qualitative results are shown in \cref{fig: grab_compare}. Further assessment of diversity against HALO is presented in \cref{fig: grab_diverse}. HALO tends to generate similar grasps for a given object. Our method produces significant diversity in terms of contact locations and grasp poses.

\input{tables/ho3d_vis}

\noindent {\bf Results on HO3D dataset.}
We evaluate the generalization capability of our models on the HO3D dataset~\cite{hampali2020honnotate}. Apart from the previously mentioned baselines, we extended our comparison to include Grasping Field~\cite{karunratanakul2020grasping} and GraspTTA~\cite{jiang2021hand}, both trained on the ObMan dataset~\cite{hasson2019learning}. As indicated in \cref{table: result_ho3d} and shown in \cref{fig: ho3d_compare}, our method achieves performance close to the best method in each metric, while keeping the highest diversity in the generated grasps. As Grasping Field~\cite{karunratanakul2020grasping} struggles to produce valid grasps for unseen objects, the computation of simulation displacement is infeasible. GraspTTA~\cite{jiang2021hand} and Grabnet~\cite{taheri2020grab} produce nearly identical grasps for a given object resulting in poor diversity. In comparison to HALO, our method achieves notably lower penetration and better stability.

\begin{figure}[t]
    \centering
     \includegraphics[width=\linewidth]{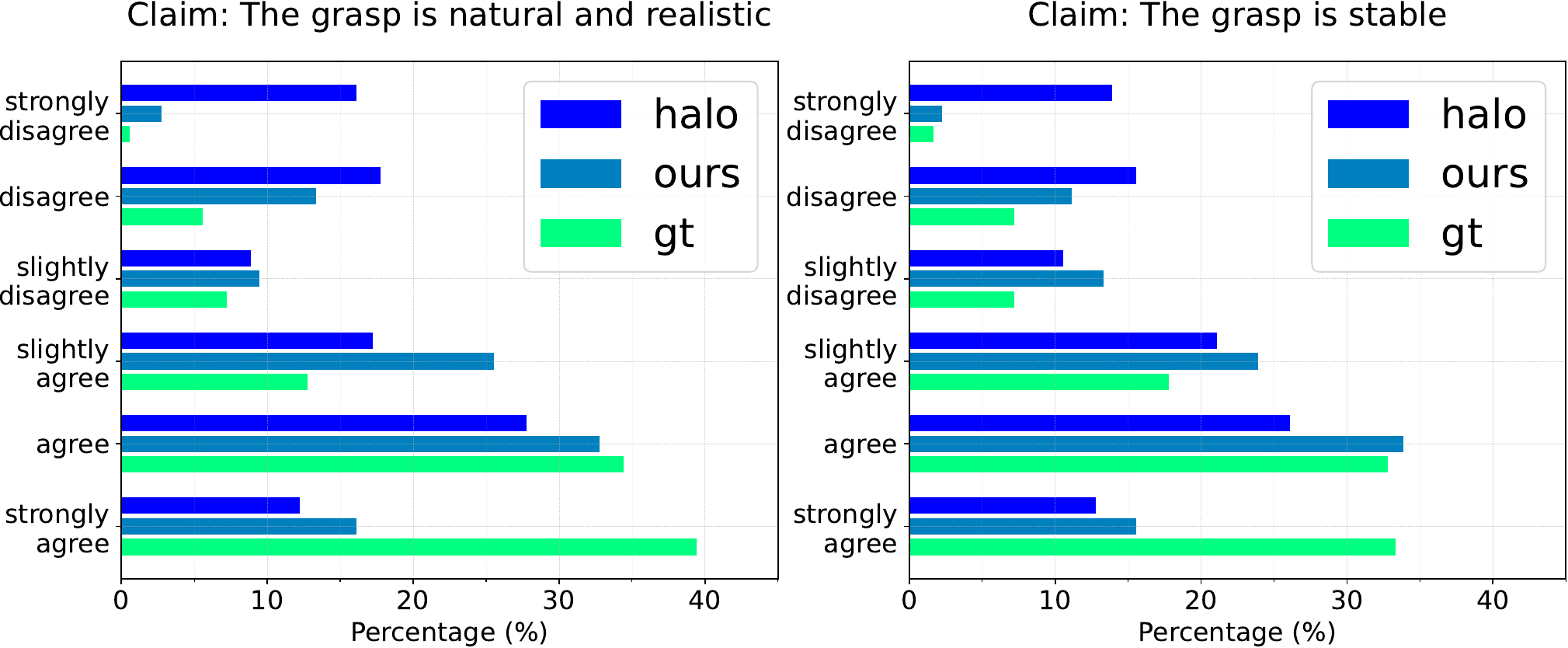} 
\caption{\textbf{Grasp human studies score distribution.} The distribution of scores shows our method achieves comparable performance to the GT in both naturalness and stability.}
\label{fig: user_study}
\end{figure}

\begin{table}[t]
\setlength{\tabcolsep}{3pt}
\renewcommand{\arraystretch}{1.05}
\footnotesize
\centering
\resizebox{\linewidth}{!}{
\begin{tabular}{lcccc}
\toprule
 & \multicolumn{2}{c}{\bf GRAB~\cite{taheri2020grab}} & \multicolumn{2}{c}{\bf HO3D~\cite{hampali2020honnotate}} \\
\cmidrule(lr){2-3} \cmidrule(lr){4-5}
\bf Method & \bf Natural $\uparrow$ & \bf Stable $\uparrow$ & \bf Natural $\uparrow$ & \bf Stable $\uparrow$\\
\midrule
HALO & 2.59 $\pm$ 1.69 & 2.68 $\pm$ 1.63 & 2.23 $\pm$ 1.41 & 2.09 $\pm$ 1.45\\
Ours & 3.21 $\pm$ 1.35 & 3.23 $\pm$ 1.23 & \textbf{3.17 $\pm$ 1.34} & \textbf{2.87 $\pm$ 1.52} \\
GT & \textbf{3.93 $\pm$ 1.18} & \textbf{3.73 $\pm$ 1.28} & - & -\\
\bottomrule
\end{tabular}}
\caption{\textbf{Grasp human study statistics.} While the gap between ours and the GT exists, our method performs better than HALO~\cite{karunratanakul2021skeleton} in terms of naturalness and stability. As HO3D dataset~\cite{hampali2020honnotate} isn't intented for grasping, GT wasn't provided.}
\label{table: user_study}
\vspace{-2mm}
\end{table}
\renewcommand{\arraystretch}{1.0}

\noindent {\bf Human evaluation.}
We also conduct a user study to assess the perceptual quality and stability of the generated grasps following~\cite{jiang2021hand, karunratanakul2020grasping, wu2022saga}. We evaluate 12 objects in total from GRAB~\cite{taheri2020grab} and HO3D~\cite{hampali2020honnotate} dataset. The evaluation involved 10 participants. For each object, we included 3 randomly sampled GT grasps from the \grab dataset, 3 generated grasps by HALO, and 3 generated grasps by ours. 
Participants were asked to rate the quality of each grasp based on its naturalness and the stability of holding the object using a five-point scale ranging from strongly disagree (0) to strongly agree (5). \cref{fig: user_study} shows the score distribution, and \cref{table: user_study} provides the mean and variance for each method's ratings. Our method outperforms HALO in terms of naturalness and stability on both datasets, but still lags behind ground truth grasps. As the HO3D dataset was not designed for grasping, appropriate ground truth hand grasps weren't available for evaluation. Further details about the human evaluation setup are available in the supplementary.

\begin{table}[t]
\setlength{\tabcolsep}{2.5pt}
\renewcommand{\arraystretch}{1.05}
\footnotesize
\vspace{1em}
\centering
\resizebox{\linewidth}{!}{
\begin{tabular}{lcccgg}
\toprule
 & \bf Penetration & \bf Contact & \bf Simulation & \bf Entropy & \bf Cluster \\
       \bf Method  &  \bf Volume $\downarrow$ & \bf Ratio $\uparrow$ & \bf Displacement $\downarrow$ & \bf $\uparrow$ & \bf Size $\uparrow$\\
\midrule
Joint & 3.40 & 0.95 & 3.29 & 2.80 & 3.87\\
Separate & 2.88 & 0.89 & 3.58 & 2.80 & \textbf{4.83} \\
Ours & \textbf{2.72} & \textbf{0.96} & \textbf{2.16} & \textbf{2.88} & 4.11 \\
\bottomrule
\end{tabular}}
\caption{\textbf{Impact of different \cgen decomposition on grasp quality.} Two ablations model the contact latent space jointly (\textbf{Joint}) or separately (\textbf{Separate}). Both approaches fail to guarantee consistency, leading to lower contact ratios, larger penetrations, and increased simulation displacements.}
\label{table:ablate_network}
\end{table}
\renewcommand{\arraystretch}{1.0}

\begin{table}[t]
\setlength{\tabcolsep}{2pt}
\renewcommand{\arraystretch}{1.05}
\footnotesize
\centering
\resizebox{\linewidth}{!}{
\begin{tabular}{l|ccc|cccgg}
\toprule
& \textbf{C} & \textbf{P} & \textbf{D} & \bf Penetration & \bf Contact & \bf Simulation & \bf Entropy & \bf Cluster \\
\bf Hand &  &  &  & \bf Volume $\downarrow$ & \bf Ratio $\uparrow$ & \bf Displacement $\downarrow$ & \bf $\uparrow$ & \bf Size $\uparrow$\\
\midrule
MANO & \cmark &  &  & 7.41 & 0.88 & 4.38 & 2.66 & 2.54 \\
MANO & \cmark  & \cmark  & \cmark  & \textbf{2.36} & \textbf{0.98} & 2.85 & 2.60 & 3.56 \\
\midrule
SDF & \cmark  &  &  & 20.13 & 0.69 & - & 2.68 & 1.44 \\
SDF & \cmark  & \cmark  &  & 2.81 & 0.66 & 6.81 & 2.80 & \textbf{5.51} \\
SDF & \cmark  & \cmark  & \cmark  & 2.72 & 0.96 & \textbf{2.16} & \textbf{2.88} & 4.11 \\
\bottomrule
\end{tabular}}
\caption{\textbf{Impact of different \cgen components on grasp quality.} Contact as \textbf{C}, part as \textbf{P}, direction as \textbf{D}. Every part of the \cgen significantly influences grasp synthesis. The proposed piecewise hand SDF model better captures fine-grained hand poses and enhances generation stability and diversity.}
\label{table: ablate_component}
\vspace{-1mm}
\end{table}
\renewcommand{\arraystretch}{1.0}

\input{tables/diversity_decomposition}

\subsection{Ablations}

\vspace{-2mm}
\noindent {\bf Contact decomposition ablations.}
We start by comparing our hierarchical \cgen decomposition with two more obvious choices: Joint and Separate modeling. Joint modeling utilizes a shared encoder to encode the three maps and a shared decoder to decode them jointly. Separate modeling encodes and decodes each \cgen component independently, using three separate encoders and decoders for each map. The results are shown in \cref{table:ablate_network}. The separate model achieves the highest diversity, and the Joint model resulted in reduced grasp diversity. But both choices struggled to maintain consistency among the three components, failing to yield physically plausible grasps. These outcomes are characterized by either larger penetrations, decreased contact ratios, or higher simulation displacements. In contrast, our proposed decomposition and sequential structure perform at its best. The generated outcomes are internally consistent and exhibit substantial diversity. 

We further illustrate the decomposition of diversity in our generated grasps across different components of the \cgen in \cref{fig: decompose_diversity}. In the top row, we present generated grasps with randomly sampled ALL three latent codes, i.e., contact, part, and direction latent codes. The second row showcases generated grasps with a fixed contact code (matching the first sample in the row) and randomly sampled the remaining two latent codes. The third-row exhibits generated grasps with fixed contact and part codes (matching the first sample in the row) and randomly sampled the direction latent codes. By conditioning on different levels of map details, we can sample and produce diverse grasps. However, as anticipated, the diversity of generated grasps decreases when we freeze more components of the \cgen representation, progressing from the top row to the bottom row. This trend becomes particularly evident in the bottom row, where the contact location and hand parts are fixed, yielding slight variations in contact directions.

\noindent {\bf Contact representation ablations.}
We conduct another ablation study to assess the contribution of different components within our \cgen representation. The study involves evaluating the generated grasp quality by removing specific components. Additionally, we compare the performance of our proposed piecewise hand SDF model against the MANO model~\cite{romero2017embodied} in contact optimization. The results are shown in \cref{table: ablate_component}. The results highlight the critical nature of all components (contact map, part map, and direction map) for achieving optimal performance. Without the guidance of the part map, both hand model struggles to generate a coherent grasp, leading to consistently higher penetrations. The influence of the part map is more critical to the piecewise hand SDF model, as it heavily relies on part information for contact reasoning. Incorporating the direction map improves contact and stability. Both the MANO model and the piecewise SDF model exhibit similar physical quality with the assistance of all three maps. Employing the SDF model better captures fine-grained hand poses, resulting in enhanced diversity and more stable outcomes.

\input{tables/twohand_vis}
\input{tables/limitation_vis}

\subsection{Synthesis Beyond Grasping}
Our proposed contact representation can extend its applications beyond grasping to address more complex \textbf{hand-hand interactions} scenarios. By substituting the object for another hand, our method, without any changes, can synthesize two-hand interactions. Specifically, we train our CVAE on a subset of the training set from \interhand dataset~\cite{moon2020interhand2}. We then use the CVAE to generate \cgen on the subset from the test split. We employ the same contact solver to decode the hand pose. By taking the left hand as input, we generate corresponding right-hand poses. The qualitative results are presented in \cref{fig: beyond_grasp}. This demonstrates the broad potential of our proposed \cgen representation in addressing a wider range of interaction tasks.

\subsection{Limitations and Future Work.} We discuss two limitations of our method. First, our approach can sometimes generate touch interactions instead of grasps. This is evident on the left side of \cref{fig: limitation}, where the generated grasp for a frying pan from the GRAB dataset~\cite{taheri2020grab} is realistic in terms of touch but not suitable for grasp. This occurs because of the inherent uncertainty in the \cgen model. These touches exhibit strong object contact but may not align well with human grasp expectations. Second, 
when our method is applied to objects outside of its training domain, like those from the HO3D dataset~\cite{hampali2020honnotate}, it occasionally creates unrealistic combinations of contact map, part map, and direction map, resulting in generated grasp with insufficient contact or significant penetration, as shown on the right side of \cref{fig: limitation}. Addressing these limitations might involve more accurately modeling the prior of the \cgen. Rather than using the Gaussian prior ($\mathcal{N} (\bm{0}, \mathbf{I})$) in VAE, advanced techniques like diffusion models could offer potential solutions. We leave further exploration of these possibilities for future work.

%% file: tables/contact_compare_vis.tex
\newcommand{\comparevis}[4]{
    \raisebox{0.5cm}{\rotatebox{90}{#4}} &
    \includegraphics[trim=#3, clip, width=0.22\linewidth]{figures/vis_contact_compare/contactopt/#1/vis_#2.png} & 
     \includegraphics[trim=#3, clip, width=0.22\linewidth]{figures/vis_contact_compare/TOCH/#1/vis_#2.png} & 
     \includegraphics[trim=#3, clip, width=0.22\linewidth]{figures/vis_contact_compare/ours/#1/vis_#2.png} & 
     \includegraphics[trim=#3, clip, width=0.22\linewidth]{figures/vis_contact_compare/gt/#1/vis_#2.png} 
    }

\begin{table}[]
    \centering
    \footnotesize
    \resizebox{\linewidth}{!}{
    \setlength{\tabcolsep}{0.2em}
    \begin{tabular}{cc|c|c|c}

       \comparevis{0}{2}{6cm 6cm 6cm 6cm}{Wineglass} \\
       \comparevis{1}{2}{6cm 6cm 6cm 6cm}{Toothpaste} \\
       \comparevis{2}{3}{2cm 3cm 5cm 5cm}{Mug} \\
       &  ContactOpt~\cite{grady2021contactopt} & TOCH~\cite{zhou2022toch} & Ours & GT \\
        
    \end{tabular}
    }

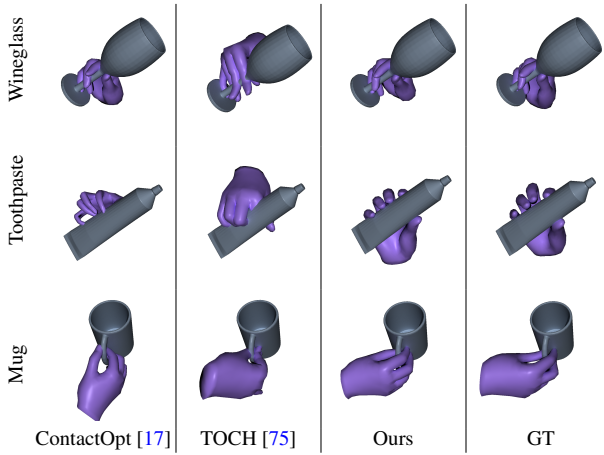
\captionof{figure}{\textbf{Contact representation comparison on \grab dataset~\cite{taheri2020grab}}. Given the object and GT contact information, we verify whether each method could recover the GT hand grasp. It can be seen both ContactOpt and TOCH exhibit failures in certain cases, whereas our method manages to achieve the closest reconstruction to the ground truth.}
\label{fig: contact_compare}
\vspace{-4mm}
\end{table}

%% file: tables/grab_vis.tex
\newcommand{\grabvis}[4]{
    \includegraphics[trim=#4, clip, width=0.12\linewidth]{figures/vis_grab_results/#1/#2/vis_#3.png}
    }

\begin{table*}[]
    \centering
    \footnotesize
    \resizebox{\linewidth}{!}{
    \setlength{\tabcolsep}{0.2em}
    \begin{tabular}{ccc|cc|cc|cc}
    
    \raisebox{0.5cm}{\rotatebox{90}{GrabNet~\cite{taheri2020grab}}} &
    \grabvis{binoculars}{grabnet}{1}{7cm 7cm 7cm 5cm} &
    \grabvis{binoculars}{grabnet}{3}{7cm 7cm 7cm 5cm} &
    \grabvis{camera}{grabnet}{1}{7cm 7cm 7cm 5cm} &
    \grabvis{camera}{grabnet}{3}{7cm 7cm 7cm 7cm} &
    \grabvis{fryingpan}{grabnet}{1}{5cm 5cm 5cm 3cm} &
    \grabvis{fryingpan}{grabnet}{3}{5cm 5cm 5cm 5cm} &
    \grabvis{mug}{grabnet}{1}{7cm 7cm 7cm 7cm} &
    \grabvis{mug}{grabnet}{3}{5cm 5cm 5cm 5cm} \\

    \raisebox{0.5cm}{\rotatebox{90}{HALO~\cite{karunratanakul2021skeleton}}} & 
    \grabvis{binoculars}{halo}{1}{7cm 7cm 7cm 5cm} &
    \grabvis{binoculars}{halo}{3}{7cm 7cm 7cm 7cm} &
    \grabvis{camera}{halo}{1}{7cm 7cm 7cm 7cm} &
    \grabvis{camera}{halo}{3}{7cm 7cm 7cm 7cm} &
    \grabvis{fryingpan}{halo}{1}{5cm 5cm 5cm 5cm} &
    \grabvis{fryingpan}{halo}{3}{5cm 5cm 5cm 5cm} &
    \grabvis{mug}{halo}{1}{7cm 7cm 7cm 5cm} &
    \grabvis{mug}{halo}{3}{5cm 5cm 5cm 5cm} \\

    \raisebox{0.5cm}{\rotatebox{90}{Ours}} & 
    \grabvis{binoculars}{ours}{1}{7cm 7cm 7cm 7cm} &
    \grabvis{binoculars}{ours}{3}{5cm 5cm 5cm 5cm} &
    \grabvis{camera}{ours}{1}{7cm 7cm 7cm 7cm} &
    \grabvis{camera}{ours}{3}{5cm 5cm 5cm 5cm} &
    \grabvis{fryingpan}{ours}{1}{5cm 5cm 5cm 5cm} &
    \grabvis{fryingpan}{ours}{3}{5cm 5cm 5cm 5cm} &
    \grabvis{mug}{ours}{1}{7cm 7cm 7cm 7cm} &
    \grabvis{mug}{ours}{3}{5cm 5cm 5cm 5cm} \\

     & \multicolumn{2}{c}{binoculars} & \multicolumn{2}{c}{camera} & \multicolumn{2}{c}{fryingpan} & \multicolumn{2}{c}{mug} \\
    \end{tabular}
    }

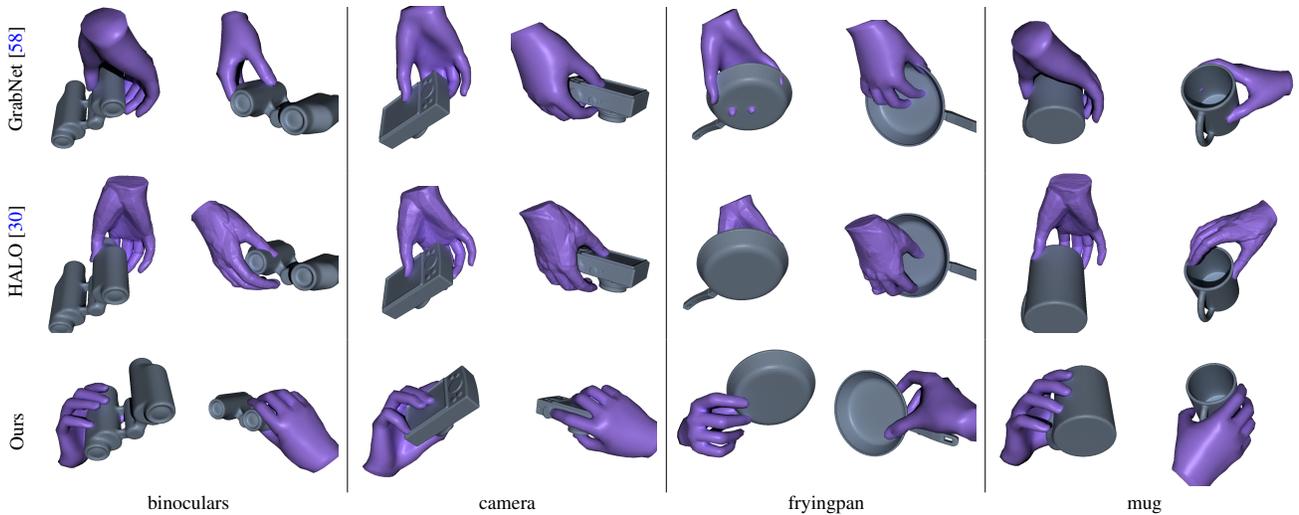
\captionof{figure}{\textbf{Qualitative comparison on \grab dataset~\cite{taheri2020grab}}. Each pair displays sampled grasps from two views. Our generated grasps showcase improved object contact and reduced penetration.} 
\label{fig: grab_compare}
\vspace{-3mm}
\end{table*}

%% file: tables/grab_diverse_vis.tex
\newcommand{\toothvis}[3]{
    \includegraphics[trim=#3, clip, width=0.12\linewidth]{figures/vis_grab_diverse/toothpaste/#1/#2/vis_1.png} &
    \includegraphics[trim=#3, clip, width=0.12\linewidth]{figures/vis_grab_diverse/toothpaste/#1/#2/vis_2.png}
    }

\newcommand{\winevis}[3]{
    \includegraphics[trim=#3, clip, width=0.12\linewidth]{figures/vis_grab_diverse/wineglass/#1/#2/vis_0.png} &
    \includegraphics[trim=#3, clip, width=0.12\linewidth]{figures/vis_grab_diverse/wineglass/#1/#2/vis_2.png}
    }

\begin{figure*}[]
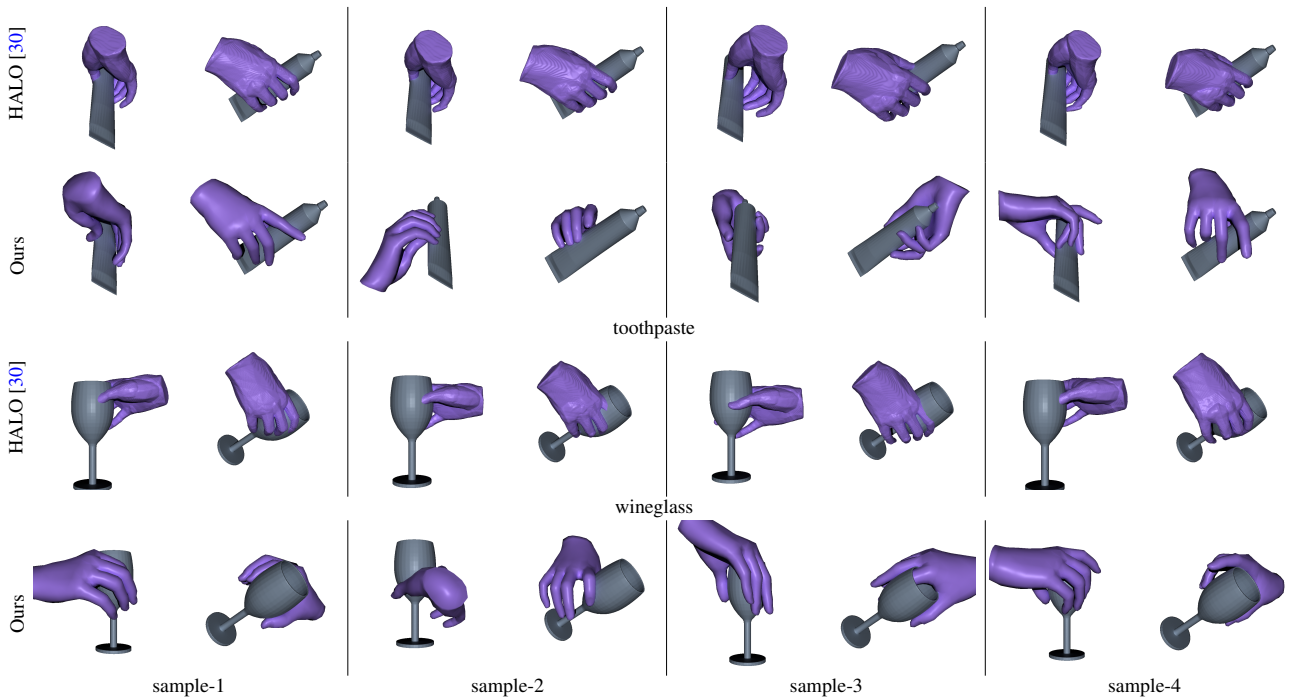


    \centering
    \footnotesize
    \resizebox{\linewidth}{!}{
    \setlength{\tabcolsep}{0.2em}
    \begin{tabular}{ccc|cc|cc|cc}

        \raisebox{0.5cm}{\rotatebox{90}{HALO~\cite{karunratanakul2021skeleton}}} &
       \toothvis{halo}{0}{5cm 5cm 5cm 5cm} &
       \toothvis{halo}{1}{5cm 5cm 5cm 5cm} &
       \toothvis{halo}{2}{5cm 5cm 5cm 5cm} &
       \toothvis{halo}{3}{5cm 5cm 5cm 5cm} \\

       \raisebox{0.5cm}{\rotatebox{90}{Ours}} &
       \toothvis{ours}{0}{5cm 5cm 5cm 5cm} &
       \toothvis{ours}{4}{5cm 5cm 5cm 5cm} &
       \toothvis{ours}{6}{5cm 5cm 5cm 5cm} &
       \toothvis{ours}{12}{5cm 5cm 5cm 5cm} \\

       \multicolumn{9}{c}{toothpaste} \\

       \raisebox{0.5cm}{\rotatebox{90}{HALO~\cite{karunratanakul2021skeleton}}} &
       \winevis{halo}{0}{5cm 5cm 5cm 5cm} &
       \winevis{halo}{1}{5cm 5cm 5cm 5cm} &
       \winevis{halo}{2}{5cm 5cm 5cm 5cm} &
       \winevis{halo}{3}{5cm 5cm 5cm 5cm} \\

       \multicolumn{9}{c}{wineglass} \\

       \raisebox{0.5cm}{\rotatebox{90}{Ours}} &
       \winevis{ours}{2}{5cm 5cm 5cm 5cm} &
       \winevis{ours}{3}{5cm 5cm 5cm 5cm} &
       \winevis{ours}{6}{5cm 5cm 5cm 5cm} &
       \winevis{ours}{14}{5cm 5cm 5cm 5cm} \\
        & \multicolumn{2}{c}{sample-1} & \multicolumn{2}{c}{sample-2} & \multicolumn{2}{c}{sample-3} & \multicolumn{2}{c}{sample-4} \\

    \end{tabular}
}
\caption{\textbf{Generated grasp diversity comparison on \grab dataset~\cite{taheri2020grab}}. Each pair displays sampled grasps from two views. We observe HALO generates similar grasps for a given input object, while ours generated grasps exhibit more diverse poses.}
\vspace{-3mm}
\label{fig: grab_diverse}
\end{figure*}

%% file: tables/ho3d_vis.tex
\newcommand{\outvis}[4]{
    \includegraphics[trim=#4, clip, width=0.12\linewidth]{figures/vis_ho3d_results/#1/#2/vis_#3.png}
    }

\begin{table*}[]
    \centering
    \footnotesize
    \resizebox{\linewidth}{!}{
    \setlength{\tabcolsep}{0.2em}
    \begin{tabular}{ccc|cc|cc|cc}

        \raisebox{0.5cm}{\rotatebox{90}{GraspTTA~\cite{jiang2021hand}}} &
        \outvis{GPMF10_0011}{grasptta}{1}{7cm 7cm 7cm 7cm} &
        \outvis{GPMF10_0011}{grasptta}{2}{7cm 7cm 7cm 7cm} &
        \outvis{GSF10_0157}{grasptta}{1}{3cm 7cm 7cm 7cm} &
        \outvis{GSF10_0157}{grasptta}{2}{7cm 7cm 7cm 7cm} &
        \outvis{MDF10_0074}{grasptta}{3}{5cm 0cm 5cm 5cm} &
        \outvis{MDF10_0074}{grasptta}{1}{5cm 5cm 5cm 5cm} &
        \outvis{SM2_0020}{grasptta}{0}{5cm 5cm 5cm 5cm} &
        \outvis{SM2_0020}{grasptta}{2}{5cm 5cm 2cm 5cm} \\

        \raisebox{0.5cm}{\rotatebox{90}{GrabNet~\cite{taheri2020grab}}} &
        \outvis{GPMF10_0011}{grabnet}{1}{7cm 7cm 7cm 7cm} &
        \outvis{GPMF10_0011}{grabnet}{2}{7cm 7cm 7cm 7cm} &
        \outvis{GSF10_0157}{grabnet}{1}{0cm 7cm 7cm 7cm} &
        \outvis{GSF10_0157}{grabnet}{2}{7cm 7cm 7cm 7cm} &
        \outvis{MDF10_0074}{grabnet}{3}{5cm 3cm 5cm 5cm} &
        \outvis{MDF10_0074}{grabnet}{1}{5cm 5cm 5cm 5cm} &
        \outvis{SM2_0020}{grabnet}{0}{5cm 5cm 5cm 5cm} &
        \outvis{SM2_0020}{grabnet}{2}{5cm 5cm 5cm 5cm} \\

        \raisebox{0.5cm}{\rotatebox{90}{HALO~\cite{karunratanakul2021skeleton}}} &
        \outvis{GPMF10_0011}{halo}{1}{7cm 7cm 7cm 7cm} &
        \outvis{GPMF10_0011}{halo}{2}{7cm 7cm 7cm 7cm} &
        \outvis{GSF10_0157}{halo}{1}{7cm 7cm 7cm 7cm} &
        \outvis{GSF10_0157}{halo}{2}{7cm 7cm 7cm 7cm} &
        \outvis{MDF10_0074}{halo}{3}{5cm 5cm 5cm 5cm} &
        \outvis{MDF10_0074}{halo}{1}{5cm 5cm 5cm 5cm} &
        \outvis{SM2_0020}{halo}{0}{5cm 5cm 5cm 5cm} &
        \outvis{SM2_0020}{halo}{2}{5cm 5cm 5cm 5cm} \\

        \raisebox{0.5cm}{\rotatebox{90}{Ours}} &
        \outvis{GPMF10_0011}{ours}{1}{7cm 7cm 7cm 7cm} &
        \outvis{GPMF10_0011}{ours}{2}{7cm 7cm 7cm 7cm} &
        \outvis{GSF10_0157}{ours}{1}{0cm 4cm 4cm 4cm} &
        \outvis{GSF10_0157}{ours}{2}{4cm 4cm 4cm 4cm} &
        \outvis{MDF10_0074}{ours}{3}{0cm 0cm 0cm 0cm} &
        \outvis{MDF10_0074}{ours}{1}{3cm 3cm 3cm 0cm} &
        \outvis{SM2_0020}{ours}{0}{0cm 5cm 5cm 5cm} &
        \outvis{SM2_0020}{ours}{2}{5cm 5cm 5cm 5cm} \\

        & \multicolumn{2}{c}{meat can} & \multicolumn{2}{c}{scissors} & \multicolumn{2}{c}{power drill} & \multicolumn{2}{c}{mustard bottle}\\
    \end{tabular}
    }
\captionof{figure}{\textbf{Qualitative comparison on out-of-domain HO3D dataset~\cite{hampali2020honnotate}}. Each pair displays sampled grasps from two views. Our method produces more plausible grasps on out-of-domain unseen objects.}
\label{fig: ho3d_compare}
\vspace{-3mm}
\end{table*}

%% file: tables/diversity_decomposition.tex
\newcommand{\decomposevis}[2]{
    \includegraphics[trim=#2, clip, width=0.24\linewidth]{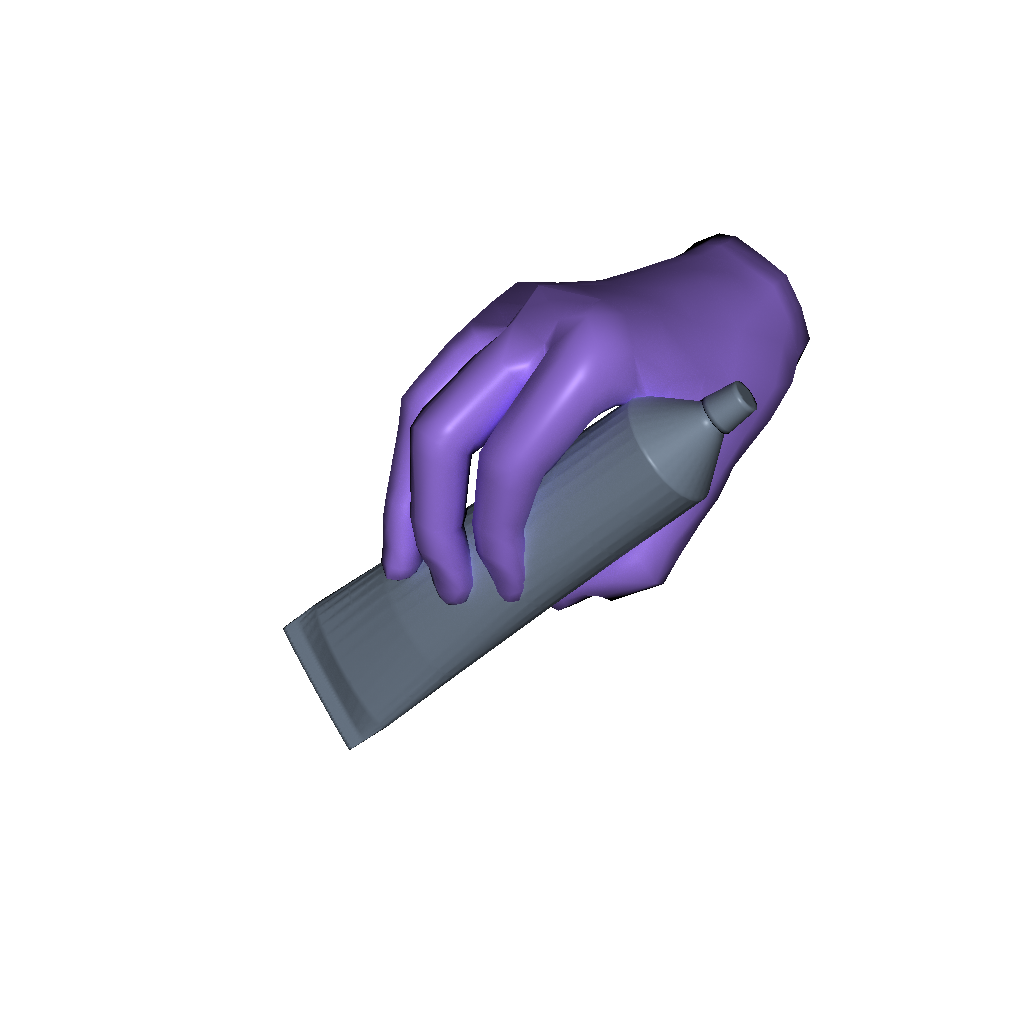} & 
    \includegraphics[trim=#2, clip, width=0.24\linewidth]{figures/diversity_decompose/#1/pred_0/vis_2.png} & 
    \includegraphics[trim=#2, clip, width=0.24\linewidth]{figures/diversity_decompose/#1/pred_1/vis_2.png} & 
    \includegraphics[trim=#2, clip, width=0.24\linewidth]{figures/diversity_decompose/#1/pred_2/vis_2.png} \\
    }

\begin{table}[]
    \centering
    \footnotesize
    \resizebox{\linewidth}{!}{
    \setlength{\tabcolsep}{0.2em}
    \begin{tabular}{cccc}

     Reference grasp & sample 1 & sample 2 & sample 3  \\
     
    \decomposevis{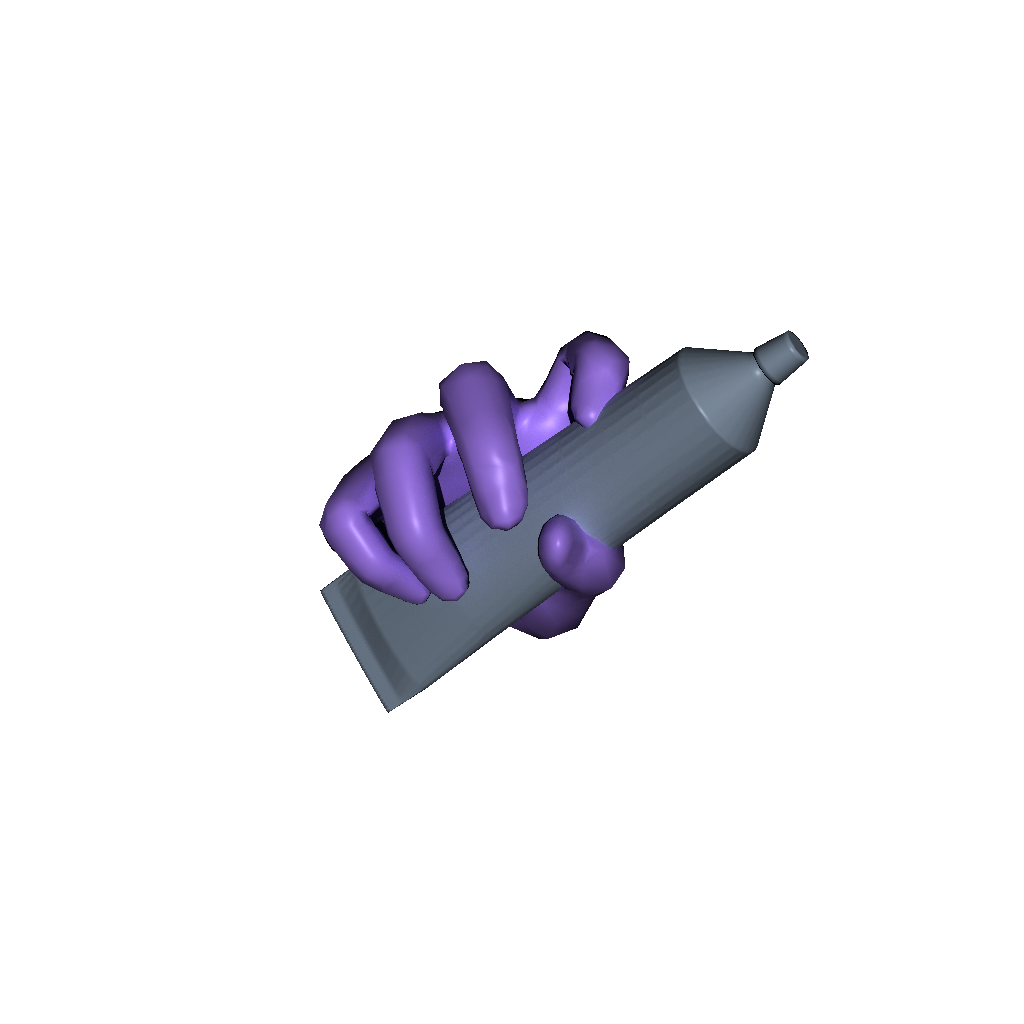}{5cm 5cm 5cm 5cm} 
     \multicolumn{4}{c}{Example 1: \textcolor{red}{Fix}: none; \textcolor{blue}{Random sample}: contact, part and direction codes} \\ \hline
    
    \decomposevis{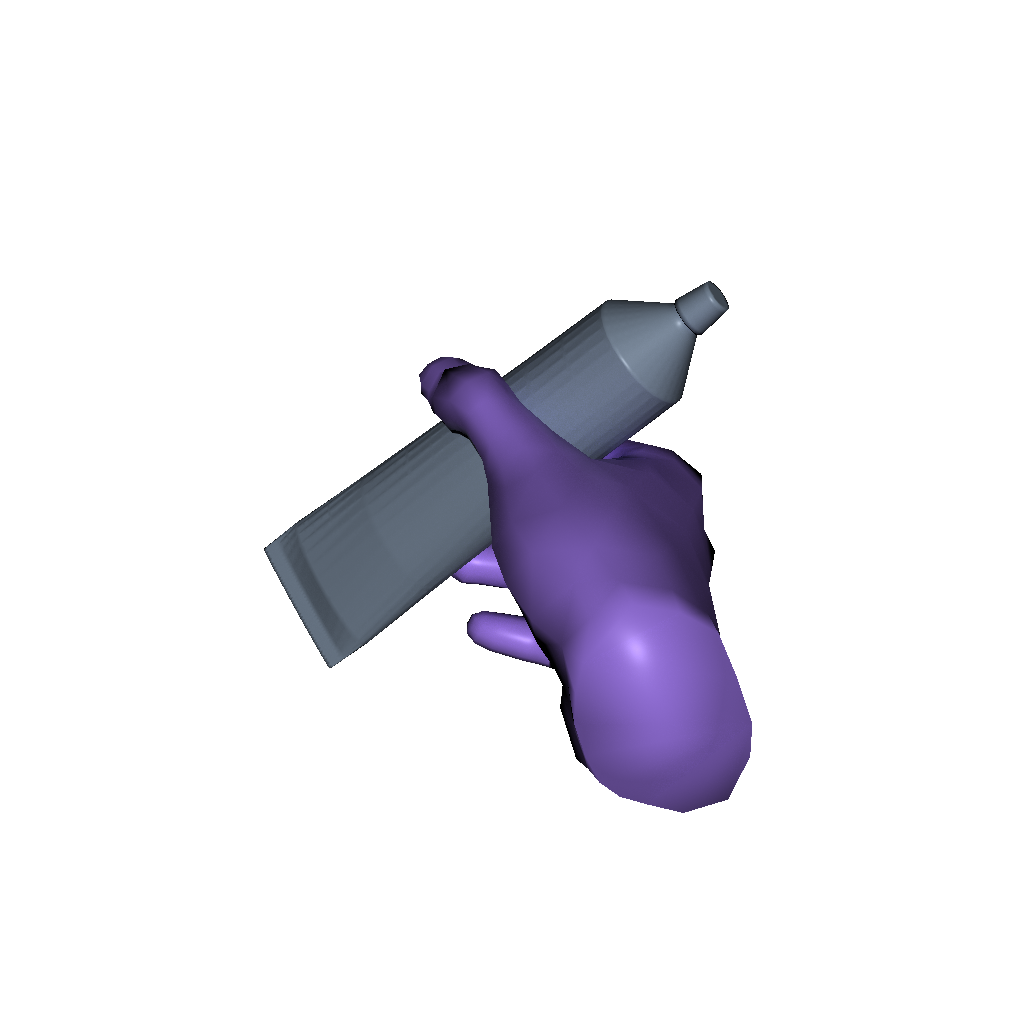}{5cm 5cm 5cm 5cm} 
    \multicolumn{4}{c}{Example 2: \textcolor{red}{Fix}: contact code; \textcolor{blue}{Random sample}: part and direction codes} \\ \hline
    
    \decomposevis{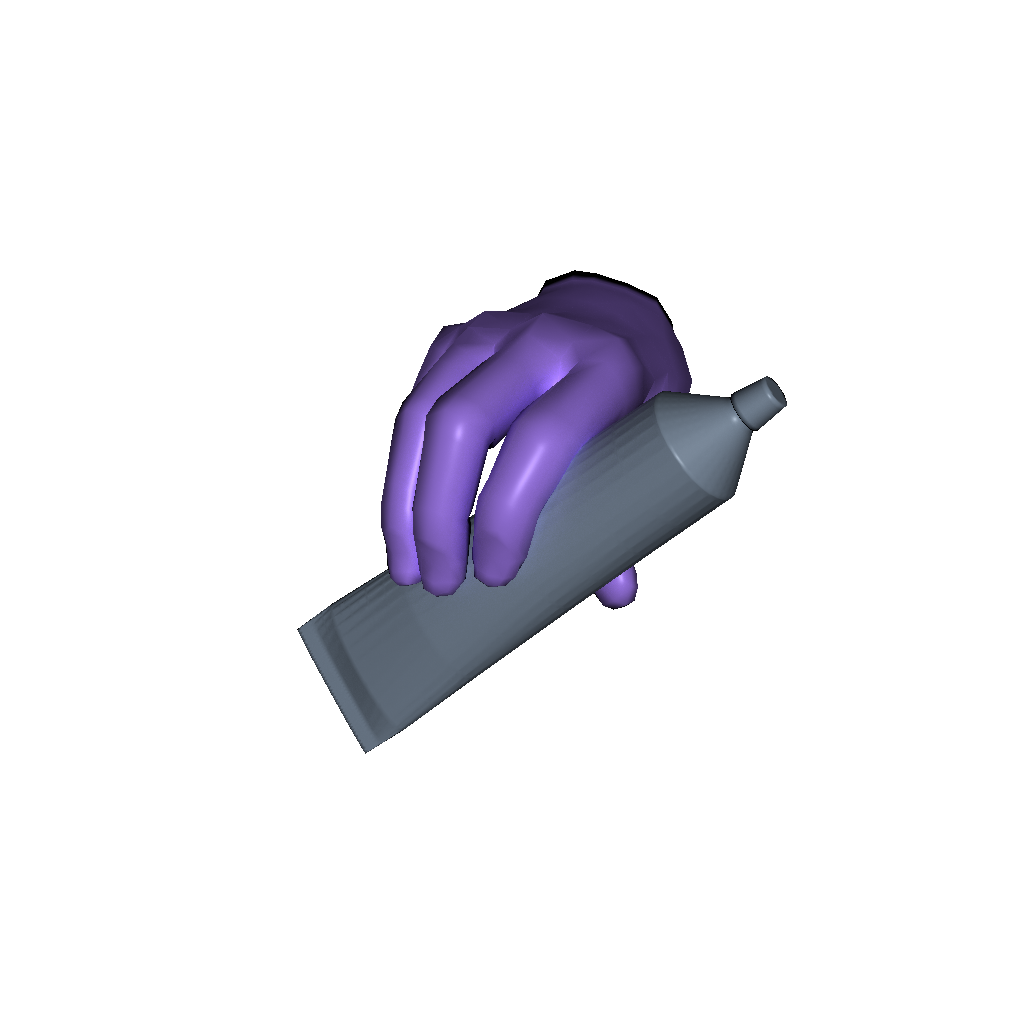}{5cm 5cm 5cm 5cm} 
     \multicolumn{4}{c}{Example 3: \textcolor{red}{Fix}: contact and part code; \textcolor{blue}{Random sample}: direction codes} \\ \hline
    
    \end{tabular}
    }
\captionof{figure}{\textbf{Visualizing grasp diversity by selectively fixing contact code, part code, and direction code.}. In the top row, grasps use random contact, part, and direction codes. In the middle row, contact is fixed, only part and direction vary. In the bottom row, contact and part are fixed, only direction changes.}
\label{fig: decompose_diversity}
\vspace{-4mm}
\end{table}

%% file: tables/twohand_vis.tex
\newcommand{\twohandvis}[3]{
    \includegraphics[trim=#3, clip, width=0.12\linewidth]{figures/vis_twohand_results/#1/vis_#2.png}
    }

\begin{table}[]
    \centering
    \footnotesize
    \resizebox{\linewidth}{!}{
    \setlength{\tabcolsep}{0.2em}
    \begin{tabular}{cc|cc}

    \twohandvis{18}{1}{7cm 7cm 7cm 7cm} & 
    \twohandvis{18}{0}{7cm 5cm 7cm 7cm} & 
    \twohandvis{58}{2}{5cm 5cm 5cm 5cm} &
    \twohandvis{58}{0}{5cm 5cm 5cm 5cm} \\
    
    \end{tabular}
    }
\captionof{figure}{\textbf{Hand-hand interaction synthesis on \interhand dataset~\cite{moon2020interhand2}.} Each pair displays a sample from two views. 
}
\label{fig: beyond_grasp}
\vspace{-6mm}
\end{table}

%% file: tables/limitation_vis.tex
\newcommand{\limitvis}[3]{
    \includegraphics[trim=#3, clip, width=0.12\linewidth]{figures/vis_limitations/#1/vis_#2.png}
    }

\begin{table}[]
    \centering
    \footnotesize
    \resizebox{\linewidth}{!}{
    \setlength{\tabcolsep}{0.2em}
    \begin{tabular}{cc|cc}

    \limitvis{grab}{2}{5cm 5cm 5cm 5cm} &
    \limitvis{grab}{3}{5cm 5cm 5cm 5cm} &
    \limitvis{ho3d}{2}{3cm 5cm 5cm 5cm} &
    \limitvis{ho3d}{3}{3cm 5cm 5cm 5cm} \\
    \end{tabular}
    }
\captionof{figure}{\textbf{Failure modes of our method.} Each pair displays a sample from two views. The \textbf{left} side shows a generated grasp appears to be more of a touch rather than a proper grasp. The \textbf{right} side shows an unsatisfactory grasp when the sampled \cgen is infeasible
on out-of-domain objects.} 
\label{fig: limitation}
\vspace{-4mm}
\end{table}

%% file: 5_conclusion.tex
\section{Conclusion}
In this work, we introduce \cgen: an object-centric contact representation for hand-object interaction. The representation is precise and complete, enabling full grasp recovery from contact information. We propose a sequential CVAE to learn the \cgen from hand-object interaction data and a model-based optimization to generate grasp from \cgen predictions of the input object. 
Experiments demonstrate our method can synthesize high-fidelity and diverse grasps for various objects. The \cgen could also be potentially used for more complex interaction scenarios synthesis beyond grasp.

%% file: 6_supp.tex
\begin{center}
\textbf{\Large Appendix}
\end{center}

\section{Interactive 3D Visualization}
High-resolution interactive qualitative results can be visualized at \href{https://stevenlsw.github.io/contactgen/}{https://stevenlsw.github.io/contactgen/}.

\section{Human Studies Setup}
\begin{figure}[htbp]
    \centering
     \includegraphics[width=\linewidth]{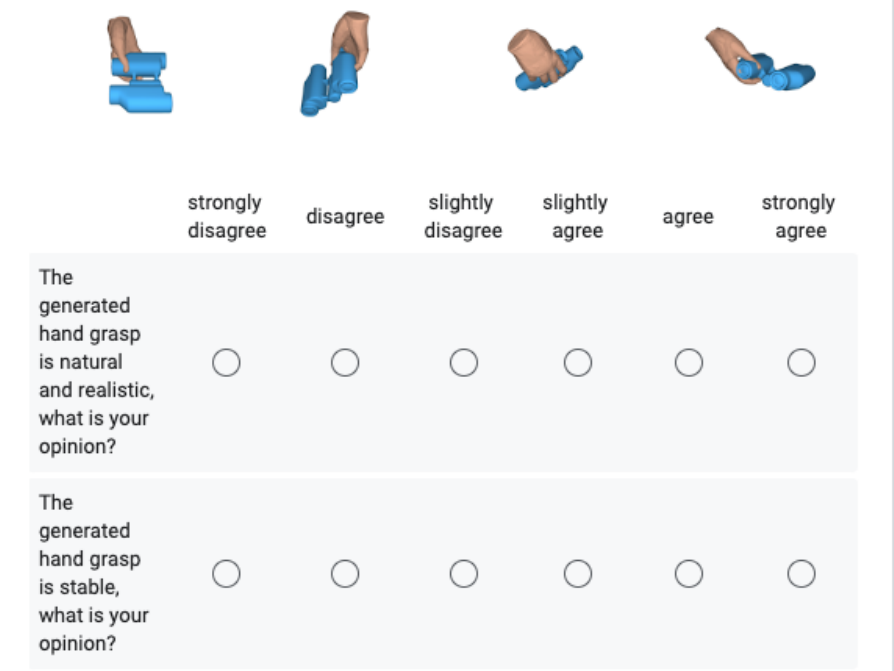} 
\caption{\textbf{Human studies interface.} Participants were asked to rate the quality of each grasp based on its naturalness and the stability using a five-point scale ranging from strongly disagree (0) to strongly agree (5).}
\label{fig: user_interface}
\vspace{-2mm}
\end{figure}

We presented 4 views for each grasp. We gathered responses from 10 participants and posed two questions for each sampled grasp: (1) The generated hand grasp is natural and realistic, what is your opinion? (2) The generated hand grasp is stable, what is your opinion? Participants rated these questions on a five-point scale, ranging from strongly disagree (0) to strongly agree (5). The human studies interface is shown in~\cref{fig: user_interface}.

\section{Implementation details of hand SDF model}
\label{sec:supp-hand-sdf}
We train the piecewise hand SDF model following~\cite{lombardi2021latenthuman}. We use the same network architecture and the same loss function as~\cite{lombardi2021latenthuman}. Each part decoder consists of four fully-connected layers with 32 neurons each, employing LeakyReLU activation with a negative slope of 0.1 for each layer. 
We use MANO shape~\cite{romero2017embodied} as the fixed global shape code shared by all part decoders. 
For each hand sample, we conducted uniform sampling of 7,000 points on the hand mesh surface, an additional 7,000 near-surface points generated by applying isotropic Gaussian noise with a mean of zero and a standard deviation of $\sigma=0.01$ to each sampled surface point, along with 1,400 randomly selected off-surface points as per Gropp et al.'s approach~\cite{gropp2020implicit}. For each on-surface sampled point, we first compute its barycentric coordinates relative to the mesh and corresponding skinning weights weighted by the neighborhood hand mesh vertices. We pick the top 2 highest skinning weights as the part label of the sampled point. The network is trained from scratch. We train it for 100 epochs with a learning rate $1e-4$ and Adam optimizer~\cite{kingma2014adam}. Once the network was trained, given the provided pose and shape code, we could compute the SDF with respect to a given query point. Subsequently, by employing the Marching Cubes algorithm~\cite{lorensen1987marching}, we could reconstruct each part under a specified pose and shape. The conversion from MANO model~\cite{romero2017embodied} to piecewise SDF model is shown in \cref{fig: hand_model_conversion}.

\begin{table}[tbhp]
    \centering
    \footnotesize
    \resizebox{\linewidth}{!}{
    \setlength{\tabcolsep}{0.2em} %
    \begin{tabular}{cc}
    
        \includegraphics[trim={10cm 10cm 10cm 10cm}, clip, width=0.48\linewidth]{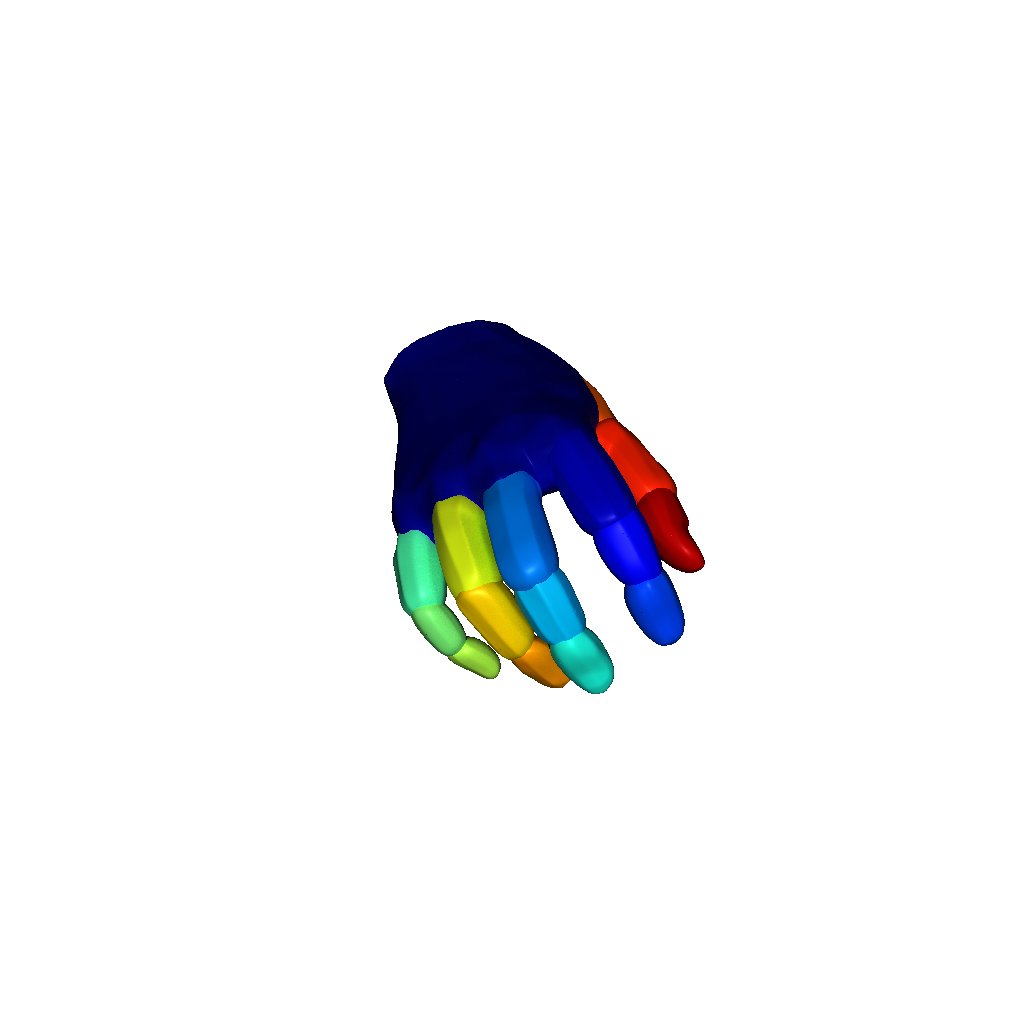} & 
        \includegraphics[trim={10cm 10cm 10cm 10cm}, clip, width=0.48\linewidth]{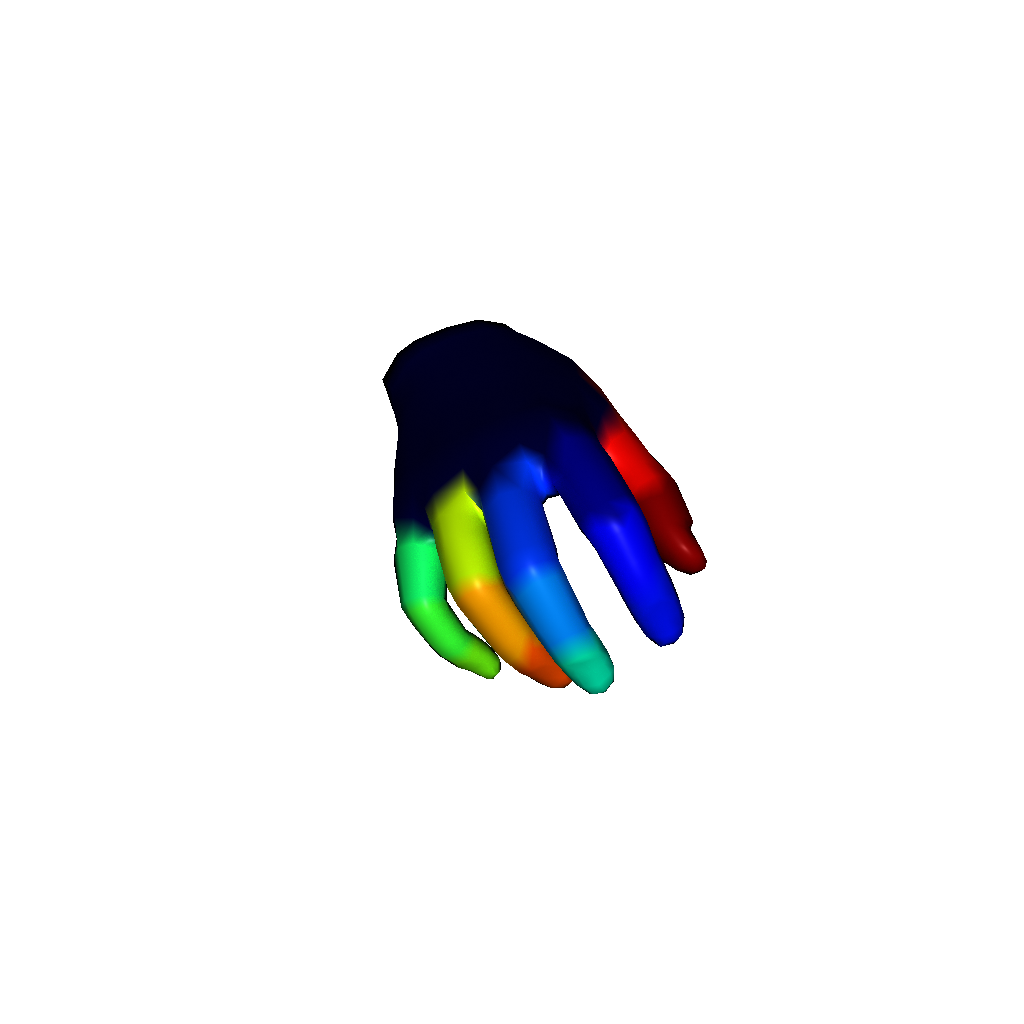} \\
    \end{tabular}
    }
\captionof{figure}{\textbf{Hand model conversion of the given pose in \cref{fig: hand_sdf}.} Left side shows the piecewise hand SDF model. Right side shows the corresponding MANO model~\cite{romero2017embodied}.}
\label{fig: hand_model_conversion}
\vspace{-5mm}
\end{table}

\section{Network architecture}
\label{sec:supp-network}
Our \cgen CVAE consists of a common backbone and three sets of encoders and decoders for each component of the \cgen. To extract features, we utilize the PointNet++~\cite{qi2017pointnet++} SSG segmentation network as the backbone. The network has three sequential set abstraction layers and three feature propagation layers: SA(512, 0.2, [64, 128]) $\rightarrow$ SA(128, 0.4, [128, 256]) $\rightarrow$ SA([256, 512]) $\rightarrow$ FP(512, 256) $\rightarrow$ FP(256, 128) $\rightarrow$ FP(128, 64). 
Each encoder is implemented as a PointNet~\cite{qi2017pointnet}, comprising a shared MLP (64, 128, 256) applied to each point's feature and max pooling across points. Pooled features are then directed to another MLP (64, 256) to generate latent distribution parameters. The MLP incorporates LeakyReLU activation with a negative slope of 0.2. The sampled latent code of each component is concatenated with per-point feature and sent to the respective decoder to infer each component map. 
The decoder architecture also employs the PointNet~\cite{qi2017pointnet} without max pooling to yield point-wise predictions of each map. To capture hand part information, we embed each part into an embedding space with a feature dimension of 64. We feed the corresponding embedded feature of the part map into the network. 
For the contact map output, we pass the decoder's output through a Sigmoid layer to normalize the result within the [0, 1] range. For the part map output, we apply the argmax operation to determine the predicted hand part label. Finally, for the direction map output, we normalize each point's output to be a unit vector.